\documentclass[10pt,journal,compsoc]{IEEEtran}
\usepackage{comm}
\usepackage{amssymb}

\usepackage{tabulary}
\usepackage{multirow}
\usepackage{amsmath}
\usepackage{color,soul}
\usepackage{siunitx}
\DeclareMathOperator*{\argmax}{argmax}

\ifCLASSOPTIONcompsoc
  \usepackage[nocompress]{cite}
\else
  \usepackage{cite}
\fi
\ifCLASSINFOpdf
   \usepackage[pdftex]{graphicx}
  \usepackage{epstopdf}
\else
\fi
\begin{document}
%
\title{Real Time Fine-Grained Categorization with Accuracy and Interpretability}

\author{Shaoli Huang~ and Dacheng Tao, ~\IEEEmembership{Fellow,~IEEE}

\IEEEcompsocitemizethanks{
\IEEEcompsocthanksitem The authors are with the Centre for Quantum Computation \& Intelligent Systems and the Faculty of Engineering and Information Technology, University of Technology, Sydney, 81 Broadway Street, Ultimo, NSW 2007, Australia. E-mail: shaol.huang@gmail.com, dacheng.tao@gmail.com.}
\thanks{Manuscript received }}

\IEEEtitleabstractindextext{
\begin{abstract}

A well-designed fine-grained categorization system usually has three contradictory requirements: accuracy (the ability to identify objects among subordinate categories); interpretability (the ability to provide human-understandable explanation of recognition system behavior); and efficiency (the speed of the system). To handle the trade-off between accuracy and interpretability, we propose a novel "Deeper Part-Stacked CNN" architecture armed with interpretability by modeling subtle differences between object parts. The proposed architecture consists of a part localization network, a two-stream classification network that simultaneously encodes object-level and part-level cues, and a feature vectors fusion component. Specifically, the part localization network is implemented by exploring a new paradigm for key point localization that first samples a small number of representable pixels and then determine their labels via a convolutional layer followed by a softmax layer. We also use a cropping layer to extract part features and propose a scale mean-max layer for feature fusion learning. Experimentally, our proposed method outperform state-of-the-art approaches both in part localization task and classification task on Caltech-UCSD Birds-200-2011. Moreover, by adopting a set of sharing strategies between the computation of multiple object parts, our single model is fairly efficient running at $32$ frames/sec.

\end{abstract}

}

\maketitle

\IEEEdisplaynontitleabstractindextext

%
\IEEEpeerreviewmaketitle

\IEEEraisesectionheading{\section{Introduction}\label{sec:introduction}}

%
%
%
%
\IEEEPARstart{F}{ine-grained} visual categorization (FGVC) refers to the task of indentifying objects from subordinate categories and is now an important subfield in object recognition. FGVC applications include, for example, recognizing species of birds \cite{welinder2010caltech,wah2011caltech,berg2014birdsnap}, pets \cite{khosla2011novel,parkhi2012cats}, flowers \cite{nilsback2008automated,angelova2013image}, and cars \cite{stark2011fine,maji2013fine}. Lay individuals tend to find it easy to quickly distinguish basic-level categories (e.g., cars or dogs), but identifying subordinate classes like "\textit{Ringed-billed gull}" or "\textit{California gull}" can be difficult, even for bird experts. Tools that aid in this regard would be of high practical value.

This task is made challenging due to the small inter-class variance caused by subtle differences between subordinaries and the large intra-class variance caused by negative factors such as differing pose, multiple views, and occlusions. However, impressive progress \cite{wah2011multiclass,berg2014birdsnap,vedaldi2014understanding,krause2015fine,xu2015augmenting} has been made over the last few years and fine-grained recognition techniques are now close to practical use in various applications such as for wildlife observation and in surveillance systems.

\begin{figure}[t]
\begin{center}
\includegraphics[width=\linewidth]{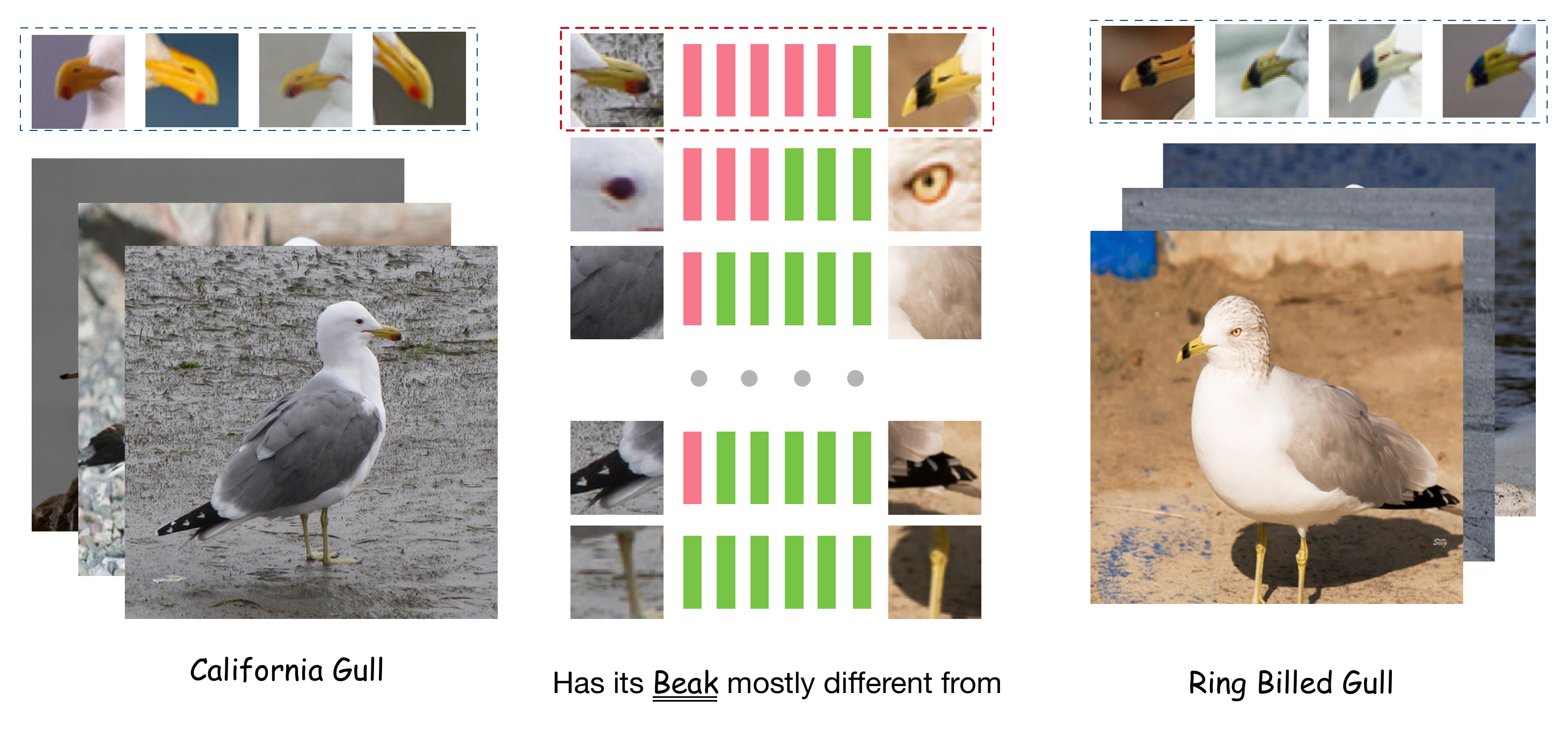}
\end{center}
   \caption{Overview of the proposed approach. We propose to classify fine-grained categories by modeling the subtle difference from specific object parts. Beyond classification results, the proposed DPS-CNN architecture also offers human-understandable instructions on how to classify highly similar object categories explicitly.}
\label{fig:title}
\end{figure}

Whilst numerous attempts have been made to boost the classification accuracy of FGVC \cite{deng2013fine,chai2013symbiotic,branson2014bird,lin2015bilinear,wang2015multiple}, an important aspect of the problem has yet to be addressed, namely the ability to generate a human-understandable "manual" on how to distinguish fine-grained categories in detail. For example, ecological protection volunteers would benefit from an algorithm that could not only accurately classify bird species but also provide brief instructions on how to distinguish very similar subspecies (a "\textit{Ringed-billed}" and "\textit{California gull}", for instance, differ only in their beak pattern, see Figure \ref{fig:title}), aided by some intuitive illustrative examples. Existing fine-grained recognition methods that aim to provide a visual field guide mostly follow a "part-based one-vs.-one features" (POOFs) \cite{berg2013poof,berg2013you,berg2014birdsnap} routine or employ human-in-the-loop methods \cite{kumar2012leafsnap,branson2014ignorant,van2015building}. However, since the amount of available data requiring interpretation is increasing drastically, a method that simultaneously implements and interprets FGVC using deep learning methods \cite{krizhevsky2012imagenet} is now both possible and advocated. 

\begin{figure*}
\begin{center}
\includegraphics[width=1.\linewidth]{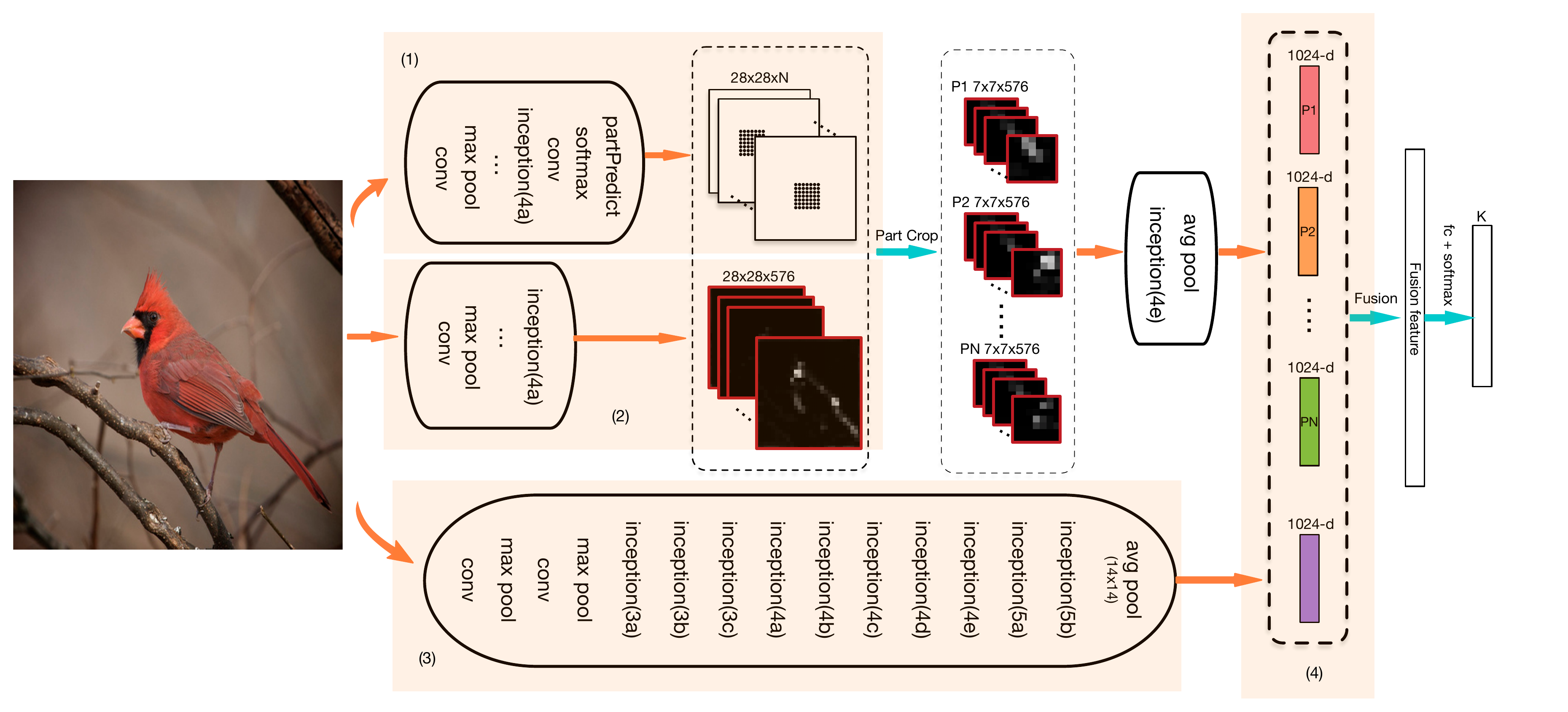}
\end{center}
   \caption{Network architecture of the proposed Deeper Part-Stacked CNN. The model consists of: (1) a fully convolutional network for part landmark localization; (2) a part stream where multiple parts share the same feature extraction procedure, while being separated by a novel part crop layer given detected part locations; (3) an object stream to capture global information; and (4) Feature fusion layer with input feature vectors from part stream and object stream to achieve the final feature representation.}
\label{fig:architecture}
\end{figure*}

It is widely acknowledged that the subtle differences between fine-grained categories mostly reside in the unique properties of object parts \cite{rosch1976basic,berg2013poof,chai2013symbiotic,maji2014part,zhang2014part,zhang2014fused}. Therefore, a practical solution to interpreting classification results as human-understandable manuals is to discover classification criteria from object parts. Some existing fine-grained datasets provide detailed part annotations including part landmarks and attributes \cite{wah2011caltech,maji2013fine}. However, they are usually associated with a large number of object parts, which incur a heavy computational burden for both part detection and classification. From this perspective, a method that follows an object part-aware strategy to provide interpretable prediction criteria at minimal computational effort but deals with large numbers of parts is desirable. In this scenario, independently training a large convolutional neural network (CNN) for each part and then combining them in a unified framework is impractical \cite{zhang2014part}. 

Here we address the fine-grained categorization problem not only in terms of accuracy and efficiency when performing subordinate-level object recognition but also with regard to the interpretable characteristics of the resulting model. We do this by learning a new part-based CNN for FGVC that models multiple object parts in a unified framework with high efficiency. Similar to previous fine-grained recognition approaches, the proposed method consists of a localization module to detect object parts (“where pathway”) and a classification module to classify fine-grained categories at the subordinate level (“what pathway”). 
In particular, our key point localization network structure is composed of a sub-network used in contemporary classification networks (AlexNet \cite{krizhevsky2012imagenet} and BN-GoogleNet \cite{ioffe2015batch}) and a 1x1 convolutional layer followed by a softmax layer to predict evidence of part locations. The inferred part locations are then fed into the classification network, in which a two-stream architecture is proposed to analyze images at both the object level (global information) and part level (local information). Multiple parts are then computed via a shared feature extraction route, separated directly on feature maps using a part cropping layer, concatenated, and then fed into a shallower network for object classification. Except for categorical predictions, our method also generates interpretable classification instructions based on object parts. Since the proposed deeper network architecture-based framework employs a sharing strategy that stacks the computation of multiple parts, we call it \emph{Deeper Part-Stacked CNN} (DPS-CNN).

\begin{figure*}
\begin{center}
\includegraphics[width=0.8\linewidth]{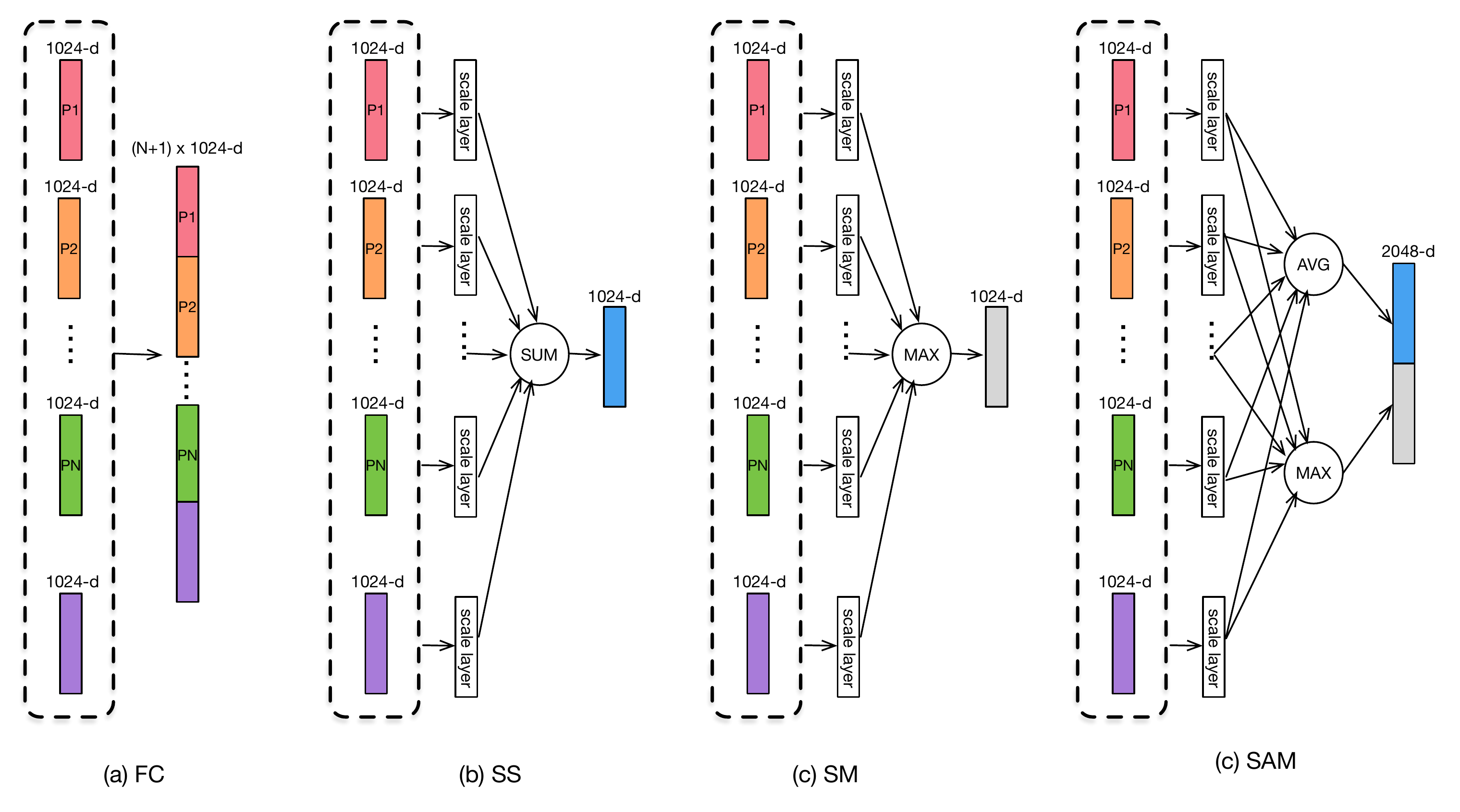}
\end{center}
   \caption{Different strategies for feature fusion which are illustrated in (a) Fully connected,(b) Scale Sum, (c) Scale Max and (d) Scale Average Max respectively.}
\label{fig:fusion}
\end{figure*}

This paper makes the following contributions: 
\begin{enumerate}
	\item DPS-CNN is the first efficient framework that not only achieves state-of-the-art performance on Caltech-UCSD Birds-200-2011 but also allows interpretation;
	\item We explore a new paradigm for key point localization, which has exceed state of the art performance on Birds-200-2011 dataset;
	\item Our classification network follows a two-stream structure that captures both object level (global) and part level (local) information, in which a new share-and-divide strategy is presented to compute multiple object parts. As a result, the proposed architecture is very efficient with a capacity of $32$ frames/sec \footnote{For reference, a single CaffeNet runs at $82$ frames/sec under the same experimental setting.} without sacrificing the fine-grained categorization accuracy. Also, we propose a new strategy called scale mean-max (SMM) for feature fusion learning. 	
\end{enumerate}

	This paper is not a direct extension of our previous work \cite{huang2016part} and several other state-of-the-art fine-grained classification models \cite{zhang2015fine,zhangspda2016,zhang2016picking, lin2015bilinear} but a significant development regarding the following aspects:
	Different to \cite{zhang2015fine} who adapts FCN for part localization, we propose a new paradigm for key point localization that first samples a small number of representable pixels and then determine their labels via a convolutional layer followed by a softmax layer; We also propose a new network architecture and enrich the methodology used in \cite{huang2016part}; Further, we introduce a simple but effective part feature encoding (named Scale Average Max) method in contrast to Bilinear in \cite{lin2015bilinear},	Spatially Weighted Fisher Vector in \cite{zhang2016picking}, and Part-based Fully Connected in \cite{zhang2016picking}.


The remainder of the paper is organized as follows. Related works are summarized in Section  \ref{sec:relatedwork}, and the proposed architecture including the localization and classification networks is described in Section  \ref{sec:dpscnn}. Detailed performance studies and analysis are presented in Section \ref{sec:exp}, and in Section \ref{sec:conclusion} we conclude and propose various applications of the proposed DPS-CNN architecture.

\section{Related Work}\label{sec:relatedwork}

\noindent\textbf{Keypoint Localization}. Subordinate categories generally share a fixed number of semantic components defined as 'parts' or 'key points' but with subtle differences in these components. Intuitively, when distinguishing between two subordinate categories, the widely accepted approach is to align components containing these fine differences. Therefore, localizing parts or key points plays a crucial role in fine-grained recognition, as demonstrated in recent works \cite{berg2013poof,zhang2014panda,maji2014part,zhang2014part,gkioxari2015actions,zhu2015deepm}. 

Seminal works in this area have relied on prior knowledge about the global shape \cite{milborrow2008locating,cootes2001active,matthews2004active,saragih2009face}. For example, the active shape model (ASM) uses a mixture of Gaussian distributions to model the shape. Although these techniques provide an effective way to locate facial landmarks, they cannot usually handle a wide range of differences such as those seen in bird species recognition. The other group of methods \cite{branson2014bird,liu2013bird,liu2014part,zhang2014part,shih2015part,lin2015deep,zhangspda2016,yu2016deep} trains a set of key point detectors to model local appearance and then uses a spatial model to capture their dependencies and has become more popular in recent years. Among them, the part localization method proposed in \cite{shih2015part,lin2015deep,zhangspda2016} is most similar to ours. In \cite{shih2015part}, a convolutional sub-network is used to predict the bounding box coordinates without using a region candidate. Although its performance is acceptable because the network is learned by jointly optimizing the part regression, classification, and alignment, all parts of the model need to be trained separately. To tackle this problem, \cite{lin2015deep} and \cite{zhangspda2016} adopt the similar pipeline of Fast R-CNN\cite{girshick2015fast}, in which part region candidates are generated to learn the part detector. In this work, we discard the common proposal-generating process and regard all receptive field centers \footnote{Here the receptive field means the area of the input image, to which a location in a higher layer feature map correspond.} of a certain intermediate layer as potential candidate key points. This strategy results in a highly efficient localization network, since we take advantage of the natural properties of CNNs to avoid the process of proposal generation. 

 \begin{figure*}
\begin{center}
\includegraphics[width=0.8\linewidth]{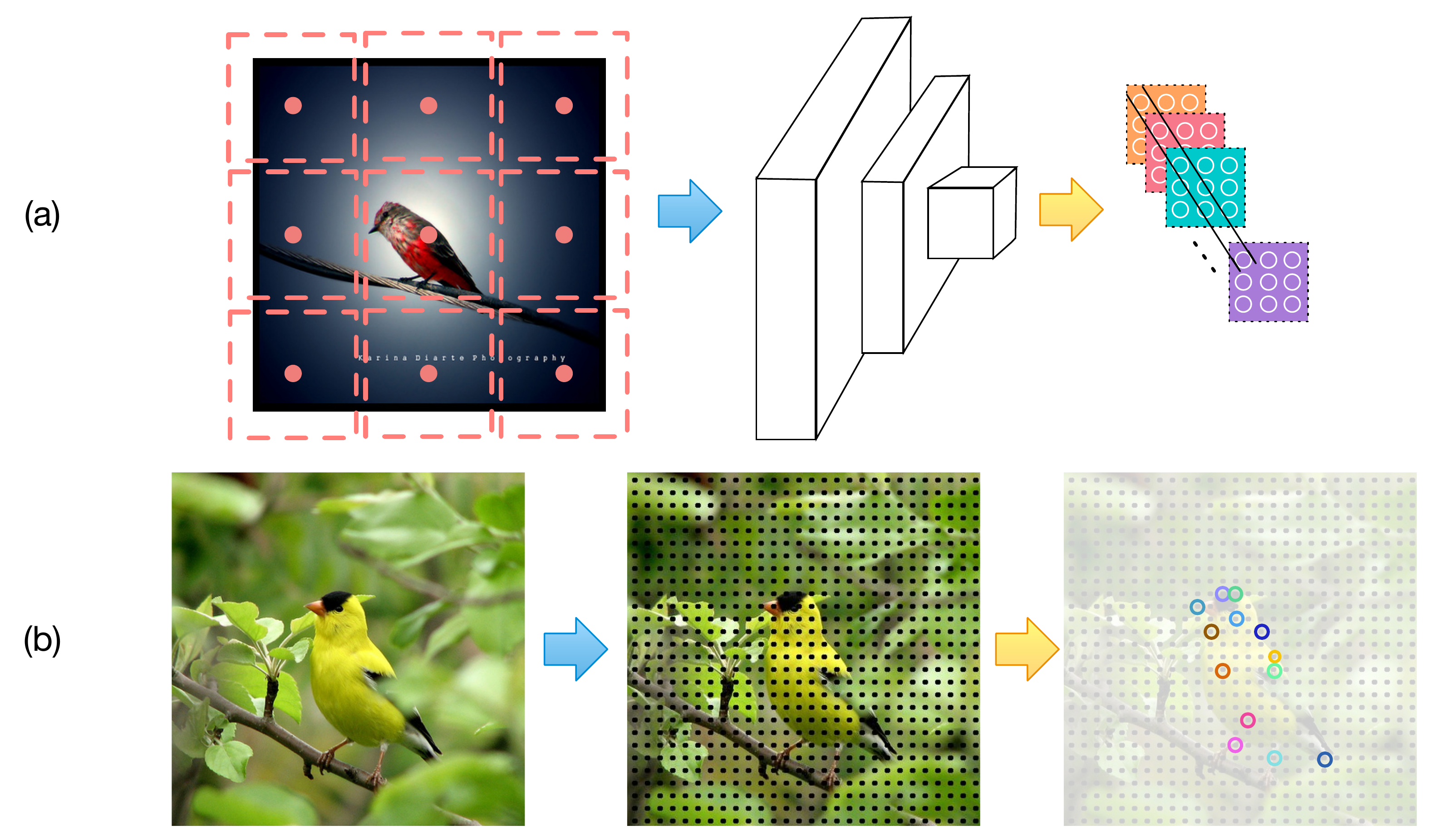}
\end{center}
   \caption{Illustration of the localization network. (a). Suppose a certain layer outputs feature maps with size 3x3 and the corresponding receptive fields are shown by dashed box. In this paper, we represent the center of each receptive filed with a feature vector at the corresponding position. (b). The first column is the input image. In the second image, each black dot is a candidate point which indicates the center of a receptive field. The final stage is to determine if a candidate point is a certain part or not. }
\label{fig:kp}
\end{figure*}

	Our work is also inspired by and inherited from fully convolutional networks (FCNs) \cite{long2015fully}, which produces dense predictions with convolutional networks. However, our network structure is best regarded as a fast and effective approach to predict sparse pixels since we only need to determine the class labels of the centers of the receptive fields of interest. Thus, FCN is more suited to segmentation, while our framework is designed for sparse key point detection. As FCN aims to predict intermediate feature maps then upsample them to match the input image size for pixel-wise prediction. Recent works \cite{zhang2015fine,wei2016convolutional} borrow this idea directly for key point localization. During training, both of these works resize the ground truths to the size of the output feature maps and then use them to supervise the network learning, while, during testing, the predicted feature maps are resized to match the input size to generate the final key point prediction. However, these methods cannot guarantee accurate position prediction due to the upsampling process.

\noindent\textbf{Fine-Grained Visual Categorization}.A number of methods have been developed to classify object categories at the subordinate level. The best performing methods have gained performance improvements by exploiting the following three aspects: more discriminative features (including deep CNNs) for better visual representation \cite{bo2010kernel,sanchez2011fisher,krizhevsky2012imagenet,szegedy2014going,simonyan2014very}; explicit alignment approaches to eliminate pose displacements \cite{branson2014bird,gavves2015local}; and part-based methods to examine the impact of object parts \cite{berg2013poof,zhang2014panda,maji2014part,zhang2014part,gkioxari2015actions,zhu2015deepm}. Another approach has been used to explore human-in-the-loop methods \cite{branson2010visual,deng2013fine,wah2014similarity} to identify the most discriminative regions for classifying fine-grained categories. Although such methods provide direct and important information about how humans perform fine-grained recognition, they are not scalable due to the need for human interactions during testing. Of these, part-based methods are thought to be most relevant to fine-grained recognition, in which the subtle differences between fine-grained categories mostly relate to the unique object part properties.

Some part-based methods \cite{berg2013poof,zhang2014part} employ strong annotations including bounding boxes, part landmarks, or attributes from existing fine-grained recognition datasets \cite{wah2011caltech,parkhi2012cats,maji2013fine,vedaldi2014understanding}. While strong supervision significantly boosts performance, the expensive human labelling process motivates the use of weakly-supervised fine-grained recognition without manually labeled part annotations, \ie, discovering object parts in an unsupervised fashion \cite{simon2015neural,krause2015fine,lin2015bilinear}.
Current state-of-the-art methods for fine-grained recognition include \cite{zhang2015fine} and \cite{lin2015bilinear}, which both employ deep feature encoding method, while DPS-CNN is largely inherited from \cite{zhang2014part}, who first detected the location of two object parts and then trained an individual CNN based on the unique properties of each part. Compared to part-based R-CNN, the proposed method is far more efficient for both detection and classification. As a result, we can use many more object parts than \cite{zhang2014part}, while still maintaining speed during testing.
 
Lin \etal \cite{lin2015bilinear}, argued that manually defined parts were sub-optimal for object recognition and thus proposed a bilinear model consisting of two streams whose roles were interchangeable as detectors or features. Although this design exploited the data-driven approach to possibly improve classification performance, it also made the resulting model difficult to interpret. In contrast, our method attempts to balance the need for classification accuracy and model interpretability in fine-grained recognition systems.




\section{Deeper Part-Stacked CNN} \label{sec:dpscnn}

A key motivation of our proposed method is to produce a fine-grained recognition system that not only considers recognition accuracy but also addresses efficiency and interpretability. To ensure that the resulting model is interpretable, we employ strong part-level annotations with the potential to provide human-understandable classification criteria. We also adapt several strategies such as sparse prediction instead of dense prediction to eliminate part proposal generation and to share computation for all part features. For the sake of classification accuracy, we learn a comprehensive representation by incorporating both global (object-level) and local (part-level) features. Based on these, in this section we present the model architecture of the proposed Deeper Part-Stacked CNN (DPS-CNN). 

According to the common framework for fine-grained recognition, the proposed architecture is decomposed into a localization network (Section \ref{subsec:localization}) and a classification network (Section \ref{subsec:classification}). In our previous work \cite{huang2016part}, we adopted CaffeNet \cite{jia2014caffe}, a slightly modified version of the standard seven-layer AlexNet architecture \cite{krizhevsky2012imagenet}, as the basic network structure. In this paper, we use a deeper but more powerful network (BN-GoogleNet) \cite{ioffe2015batch} as a substitute. A unique feature of our architecture is that the message transferring operation from the localization network to the classification network, which uses the detected part locations to perform part-based classification, is conducted directly on the \textit{Inception-4a} output feature maps within the data forwarding process. This is a significant departure from the standard two-stage pipeline of part-based R-CNN, which consecutively localizes object parts and then trains part-specific CNNs on the detected regions. Based on this design, sharing schemes are performed to make the proposed DPS-CNN fairly efficient for both learning and inference. Figure \ref{fig:architecture} illustrates the overall network architecture.

\subsection{Localization Network} \label{subsec:localization}

The first stage in our proposed architecture is a localization network that aims to detect the location of object parts. We employ the simplest form of part landmark annotation, where a 2D key point is annotated at the center of each object part. Assume that $M$ - the number of object parts labeled in the dataset is sufficiently large to offer a complete set of object parts in which fine-grained categories are usually different. A naive approach to predicting these key points is to directly apply FCN architecture \cite{long2015fully} for dense pixel-wise prediction. However, this method usually biases the learned predictor because, in this task and unlike semantic segmentation, the number of key point annotations is extremely small compared to the number of irrelevant pixels.

\begin{figure*}
\begin{center}
\includegraphics[width=\linewidth]{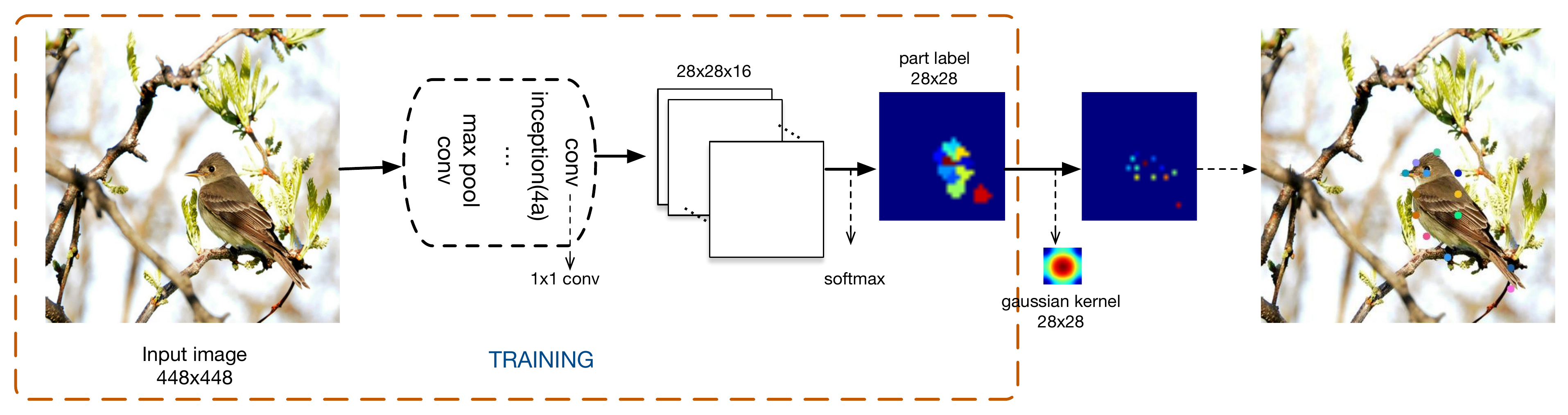}
\end{center}
   \caption{Demonstration of the localization network. Training process is denoted inside the dashed box. For inference, a Gaussian kernel is then introduced to remove noise. The results are $M$ 2D part locations in the $27\times27$ \textit{conv5} feature map.}
\label{fig:fcn}
\end{figure*}

Motivated by the recent progress in object detection \cite{ren2015faster} and semantic segmentation \cite{long2015fully}, we propose to use the centers of receptive fields as key point candidates and use a fully convolutional network to perform sparse pixel prediction to locate the key points of object parts (see Figure \ref{fig:kp}(b)). In the field of object detection, box candidates expected to be likely objects are first extracted using proposal-generating methods such as selective search \cite{uijlings2013selective} and region proposal networks \cite{ren2015faster}. Then, CNN features are learned to represent these box candidates and finally used to determine their class label. We adapt this pipeline to key point localization but omit the candidate generation process and simply treat the centers of receptive fields corresponding to a certain layer as candidate points. As shown in Figure \ref{fig:kp}(a), the advantage of using this method is that each candidate point can be represented by a $1D$ cross-channel feature vector in the output feature maps. Also, in our candidate point evaluation experiments in Table \ref{tab:recall}, we find that given an input image of size $448$x$448$ and using the receptive fields of the \textit{inception-4a} layer in BN-GoogleNet generates $28$x$28$ candidate points and $100$\% recall at PCK@$0.1$.
 \\

\noindent\textbf{Fully convolutional network.} An FCN is achieved by replacing the parameter-rich fully connected layers in standard CNN architectures constructed by convolutional layers with kernels of spatial size $1\times1$. Given an input RGB image, the output of an FCN is a feature map of reduced dimension compared to the input. The computation of each unit in the \textit{feature map} only corresponds to pixels inside a region of fixed size in the input image, which is called its \textit{feature map}. We prefer FCNs because of the following reasons: (1) feature maps generated by FCNs can be directly utilized as the part locating results in the classification network, as detailed in Section \ref{subsec:classification}; (2) the results of multiple object parts can be obtained simultaneously; (3) FCNs are very efficient for both learning and inference. \\


\noindent\textbf{Learning.} We model the part localization process as a multi-class classification problem on sparse output spatial positions. Specifically, suppose the output of the last FCN convolutional layer is of size $h\times w\times d$, where $h$ and $w$ are spatial dimensions and $d$ is the number of channels. We set $d=M+1$. Here, $M$ is the number of object parts and $1$ denotes an additional channel to model the background. To generate corresponding ground-truth labels in the form of feature maps, units indexed by $h \times w$ spatial positions are labeled with their nearest object part; units that are not close to any of the labeled parts (with an overlap with respect to a receptive field) are labeled as background. In this way, ground-truth part annotations are transformed into the form of corresponding feature maps, while in recent works that directly apply FCNs \cite{zhang2015fine,wei2016convolutional}, the supervision information is generated by directly resizing the part ground-truth image. 

Another practical problem here is determining the model depth and the input image size for training the FCN. Generally, layers at later stages carry more discriminative power and, therefore, are more likely to generate good localization results; however, their receptive fields are also much larger than those of previous layers. For example, the receptive field of the \textit{inception-4a} layer in BN-GoogleNet has a size of $107\times107$ compared to the $224\times224$ input image, which is too large to model an object part. We propose a simple trick to deal with this problem, namely upsampling the input images so that the fixed size receptive fields denoting object parts become relatively smaller compared to the whole object, while still using later stage layers to guarantee discriminative power. In the proposed architecture, the input image is upsampled to double the resolution and the \textit{inception-4a} layer is adopted to guarantee discrimination. 

The localization network is illustrated in Figure \ref{fig:fcn}. The input images are warped and resized into a fixed size of $448\times448$. All layers from the beginning to the \textit{inception-4a} layer are cut from the BN-GoogleNet architecture, so the output size of the \textit{inception-4a} layer is $28\times28\times576$. Then, we further introduce an $1\times1$ convolutional layer with $M+1$ outputs termed \textit{conv} for classification. By adopting a location-preserving softmax that normalizes predictions at each spatial location of the feature map, the final loss function is a sum of softmax loss at all $28\times28$ positions:

\begin{equation}\label{eqn:fcnloss}
L = -\sum_{h=1}^{28}\sum_{w=1}^{28} \log\sigma(h,w,\hat{c}),
\end{equation}
where
\begin{displaymath}
\sigma(h,w,\hat{c}) = \frac{\exp(f_{conv}(h,w,\hat{c}))}{\sum_{c=0}^M \exp(f_{conv}(h,w,c))}.
\end{displaymath}
Here, $\hat{c}\in[0,1,...,M]$ is the part label of the patch at location $(h,w)$, where the label $0$ denotes background. $f_{conv}(h,w,c)$ stands for the output of \textit{conv} layer at spatial position $(h,w)$ and channel $c$.
\\

\noindent\textbf{Inference.} Inference starts from the output of the learned FCN, \ie, $(M+1)$ part-specific heat maps of size $28\times28$, in which we introduce a Gaussian kernel $\mathcal{G}$ to remove isolated noise in the feature maps. The final output of the localization network are $M$ locations in the $28\times28$ \textit{conv} feature map, each of which is computed as the location with the maximum response for one object part.

Meanwhile, considering that object parts may be missing in some images due to varied poses and occlusion, we set a threshold $\mu$ that if the maximum response of a part is below $\mu$, we simply discard this part's channel in the classification network for this image. Let $g(h,w,c)=\sigma(h,w,c)*\mathcal{G}$, the inferred part locations are given as:

\begin{equation}
(h_c^*,w_c^*)=
  \left\{
   \begin{array}{ll}
   \argmax_{h,w} g(h,w,c) & \text{if } g(h_c^*,w_c^*,c)>\mu, \\
   (-1,-1) & \text{otherwise.}
   \end{array}
  \right.
\end{equation}

\subsection{Classification network}\label{subsec:classification}
The second stage of the proposed DPS-CNN is a classification network with the inferred part locations given as an input. As shown in Figure \ref{fig:architecture}, it follows a two-stream architecture with a \emph{Part Stream} and a \emph{Object Stream} to capture semantics from different angles. The outputs of both two streams are fed into a feature fusion layer followed by a fully connected layer and a softmax layer.\\

\noindent\textbf{Part stream.}
The part stream is the core of the proposed DPS-CNN architecture. To capture object part-dependent differences between fine-grained categories, one can train a set of part CNNs, each one of which conducts classification on a part separately, as proposed by Zhang \etal \cite{zhang2014part}. Although such method works well for situations employing two object parts \cite{zhang2014part}, we argue that this approach is not applicable when the number of object parts is much larger, as in our case, because of the high time and space complexities.

We introduce two strategies to improve part stream efficiency, the first being model parameter sharing. Specifically, model parameters of layers before the part crop layer and \textit{inception-4e} are shared among all object parts and can be regarded as a generic part-level feature extractor. This strategy reduces the number of parameters in the proposed architecture and thus reduces the risk of overfitting. 
We also introduce a \emph{part crop} layer as a computational sharing strategy. The layer ensures that the feature extraction procedure of all parts only requires one pass through the convolutional layers. 

After performing the shared feature extraction procedure, the computation of each object part is then partitioned through a \emph{part crop} layer to model part-specific classification cues. As shown in Figure \ref{fig:architecture}, the input for the part crop layer is a set of feature maps (the output of \textit{inception-4a} layer in our architecture) and the predicted part locations from the previous localization network, which also reside in \textit{inception-4a} feature maps. For each part, the part crop layer extracts a local neighborhood centered on the detected part location. Features outside the cropped region are simply discarded. In practice, we crop $l\times h$ neighborhood regions from the $28\times 28$ \textit{inception-4a} feature maps. The cropped size of feature regions may have an impact on recognition performance, because larger crops will result in redundancy when extracting multiple part features, while smaller crops cannot guarantee rich information. For simplicity, we use $l=h=7$ in this paper to ensure that the resulting receptive field is large enough to cover the entire part.\\

\noindent\textbf{Object stream.}
The object stream captures object-level semantics for fine-grained recognition. It follows the general architecture of BN-GoogleNet, in which the input of the network is a $448\times 448$ RGB image and the output of \textit{incenption-5b} layer are $14\times 14$ feature maps. Therefore, we use $14\times 14$ average pooling instead of $7 \times 7$ in original setting. 


The design of the two-stream architecture in DPS-CNN is analogous to the famous Deformable Part-based Models \cite{felzenszwalb2010object}, in which object-level features are captured through a root filter in a coarser scale, while detailed part-level information is modeled by several part filters at a finer scale. We find it critical to measure visual cues from multiple semantic levels in an object recognition algorithm.

We conduct the standard gradient descent to train the classification network. It should be noted, however, that the gradient of each element $\frac{\partial E}{\partial X_{i,j}}$ in \textit{inception-4a} feature maps is calculated by the following equation:
\begin{equation}
\frac{\partial E}{\partial X_{i,j}} = \sum_{c=1}^M \phi (\frac{\partial E}{\partial X^c_{i,j}}),
\end{equation}
where $E$ is the loss function, $X^c_{i,j}$ is the feature maps cropped by part $c$ and
\begin{equation}
\phi (\frac{\partial E}{\partial X^c_{i,j}})=
\begin{cases}
\frac{\partial E}{\partial X^c_{i,j}} &\mbox{$X_{i,j}$ corresponding to  $X^c_{i,j}$},  \\
0 &\mbox{otherwise}.
\end{cases}
\end{equation}

Specifically, the gradient of each cropped part feature map (in $7\times 7$ spatial resolution) is projected back to the original size of \textit{inception-4a} ($28\times28$ feature maps) according to the respective part location and then summed.
The computation of all other layers simply follows the standard gradient rules.
Note that the proposed DPS-CNN is implemented as a two stage framework, \ie after training the FCN, weights of the localization network are fixed when training the classification network. \\

\noindent\textbf{Feature Fusion}

The commonest method \cite{lin2015deep,zhang2014part} for combining all part-level and object-level features is to simply concatenate all these feature vectors as illustrated in Figure \ref{fig:fusion}(a). However, this approach may cause feature redundancy and also suffer from high-dimensionality when part numbers become large. 
To effectively utilize all part- and object-level features, we present three options for learning fusion features: scale sum (SS), scale max (SM), and scale mean-max (SMM), as illustrated in Figure \ref{fig:fusion}(a), Figure \ref{fig:fusion}(b), and Figure \ref{fig:fusion}(d), respectively. All three methods include the shared process of placing a scale layer on top of each branch. Nevertheless, as indicated by their names, the scale sum feature is the element-wise sum of all output branches, the scale max feature is generated by an element-wise maximum operation, while the scale average-max feature is the concatenation of element-wise mean and max features. 
In our previous work \cite{huang2016part} based on the standard CaffeNet architecture, each branch from the part stream and the object stream was connected with an independent \emph{fc6} layer to encourage diversity features, and the final fusion feature was the sum of all the outputs of these \emph{fc6} layers. As this fusion process requires $M+1$ times model parameters more than the original \emph{fc6} layer in CaffeNet and consequently incurs a huge memory cost, a $1\times1$ convolutional layer is used for dimensionality reduction. 
Here we redesign this component for simplicity and to improve performance. First, a shared inception module is placed on top of the cropped part region to generate higher level features. Also, a scale layer follows each branch feature to encourage diversity between parts. Furthermore, the scale layer has fewer parameters than the fully connected layer and, therefore, reduces the risk of overfitting and decreases the model storage requirements.

\begin{table*}
\centering
\caption{Localization recall of candidate points selected by \textit{inception-4a} layer with different $\alpha$ values. The abbreviated part names from left to right are: Back, Beak, Belly, Breast, Crown, Forehead,Left Eye,Left Leg, Left Wing, Nape, Right Eye, Right Leg, Right Wing, Tail, and Throat}
\label{tab:recall}       

\begin{tabulary}{17.5cm}{cccccccccccccccccc}

\hline\noalign{\smallskip}

$\alpha$ & Ba & Bk & Be & Br & Cr & Fh & Le & Ll & Lw & Na & Re & Rl & Rw & Ta & Th & Avg \\
\noalign{\smallskip}\hline\noalign{\smallskip}
$0.05$ & 100 & 100 & 100 & 100 & 100 & 100 & 100 & 100 & 100 & 100 & 100 & 100 & 100 & 100 & 100 & 100  \\

$0.02$ & 90.8 & 89.8 & 90.8 & 90.4 & 90.9 & 91.4 & 90.4 & 90.4 & 90.0 & 90.7 & 90.3 & 89.9 & 90.3 & 90.5 & 90.3 & 90.5 \\

$0.01$ & 26.8 & 26.3 & 9.1 & 11.2 & 5.2 & 4.1 & 40.4 & 9.4 & 10.8 & 14.6 & 9.9 & 11.9 & 9.6 & 11.2 & 22.3 & 13.2  \\
\noalign{\smallskip}\hline

\end{tabulary}
\end{table*}

\section{Experiments}\label{sec:exp}
In this section we present experimental results and a thorough analysis of the proposed method. Specifically, we evaluate the performance from four different aspects: localization accuracy, classification accuracy, inference efficiency, and model interpretation.

\begin{figure*}[!t]
\begin{center}
\includegraphics[width=\linewidth]{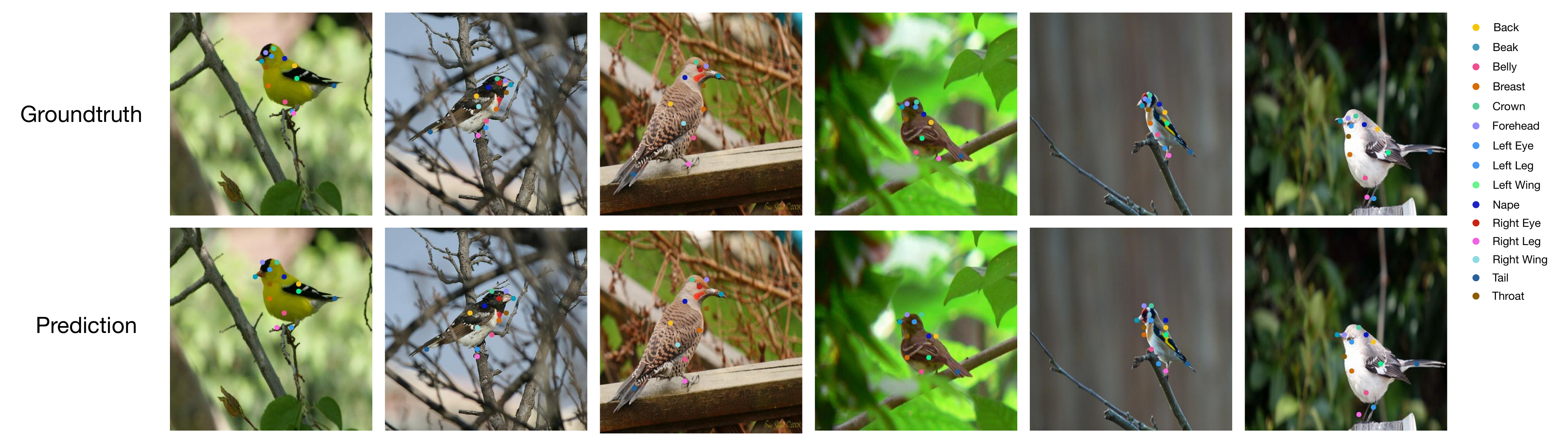}
\end{center}
   \caption{Typical localization results on CUB-200-2011 test set. Better viewed in color.}
\label{fig:loc}
\end{figure*}

\subsection{Dataset and implementation details}
Experiments are conducted on the widely used fine-grained classification benchmark the Caltech-UCSD Birds dataset (CUB-200-2011) \cite{wah2011caltech}. The dataset contains $200$ bird categories with roughly $30$ training images per category. In the training phase we adopt strong supervision available in the dataset, \ie we employ 2D key point part annotations of altogether $M=15$ object parts together with image-level labels and object bounding boxes.

The labeled parts\footnote{The $15$ object parts are back, beak, belly, breast, crown, forehead, left eye, left leg, left wing, nape, right eye, right leg, right wing, tail, and throat.} imply places where people usually focus on when being asked to classify fine-grained categories; thus they provide valuable information for generating human-understandable systems.

The proposed Deeper Part-Stacked CNN architecture is implemented using the open-source package Caffe \cite{jia2014caffe}.
Specifically, input images are warped to a fixed size of $512\times512$, randomly cropped into $448\times448$, and then fed into the localization network and the part stream in the classification network as input. We employ a pooling layer with kernel $7\times7$ to guarantee synchronization between the two streams in the classification network.


\subsection{Candidate keypoints}

For the key point localization task, we follow the proposal-based object detection method pipeline; centers of receptive fields corresponding to a certain layer are first regarded as candidate points and then forwarded to a fully convolutional network for further classification. Similar to object detection using proposals, whether selected candidate points have a good coverage of pixels of interest in the test image plays a crucial role in key point localization, since missed key points cannot be recovered in subsequent classification. Thus, we first evaluate the candidate point sampling method. The evaluation is based on the PCK metric \cite{yang2013articulated}, in which the error tolerance is normalized with respect to the input image size. For consistency with evaluation of key point localization, a ground truth point is recalled if there exists a candidate point matched in terms of the PCK metric. Table \ref{tab:recall} shows the localization recall of candidate points selected by \textit{inception-4a} with different $\alpha$ values $0.05$, $0.02$ and $0.01$. 
	 As expected, candidate points sampled by layer \textit{inception-4a} have a great coverage of ground truth using PCK metric with $\alpha > 0.02$. However, the recall drop dramatically when using $\alpha = 0.01$. This mainly because of the large stride(16) in \textit{inception-4a} layer, which results in the distance between two closest candidate points is 16 pixels, while setting a input size of $448$ with $\alpha = 0.01$ requires the candidate point should be close to the ground truth within $4.48$ pixels.

\begin{figure*}
\begin{center}
\includegraphics[width=1\linewidth]{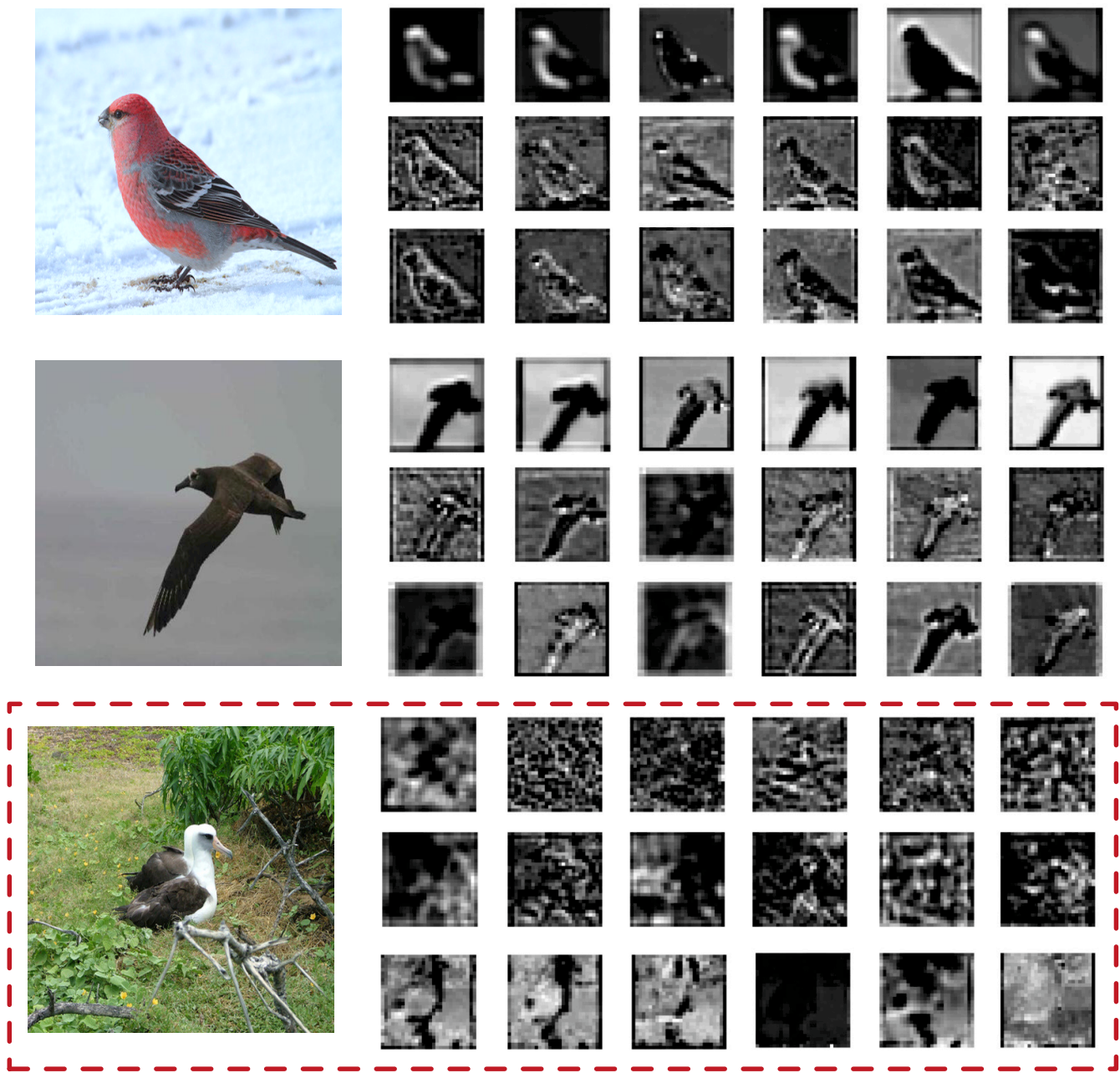}
\end{center}
   \caption{Feature maps visualization of \emph{Inception-4a} layer. Each example image is followed by three rows of top six scoring feature maps, which are from the part stream, object stream and and baseline BN-inception network respectively. Red dash box indicates a failure case of visualization using the model learned by our approach.}
\label{fig:fm}
\end{figure*}

\subsection{Localization Results}

Following \cite{long2014convnets}, we consider a key point to be correctly predicted if the prediction lies within a Euclidean distance of $\alpha$  times the maximum of the input width and height compared to the ground truth. Localization results are reported on multiple values of $\alpha\in \{0.1, 0.05, 0.02\}$ in the analysis below. The value  $\alpha$ in the PCK metric is introduced to measure the error tolerance in key point localization. 
To investigate the effect of the selected layer for key point localization, we perform experiments using the \textit{inception-4a},\textit{inception-4b},\textit{inception-4c} and \textit{inception-4d} layers as part detector layers. As shown in Table \ref{tab:loc}, a higher layer with a larger receptive field tends to achieve better localization performance than a lower layer with $\alpha=0.1$. This is mainly because the larger receptive fields are crucial for capturing spatial relationships between parts and improve performance (see Table \ref{tab:rf}). However, in contrast, for $\alpha = 0.05$ or $0.02$, the performance decreases at deeper layers. One possible explanation is that although higher layers obtain better semantic information about the object, they lose more detailed spatial information. To evaluate the effectiveness of our key point localization approach, we also compare it with recent published works \cite{huang2016part, zhang2015fine,yu2016deep} providing PCK evaluation results on CUB-200-2011 along with experimental results using a more reasonable evaluation metric called average precision of key points (APK), which correctly penalizes both missed and false-positive detections \cite{yang2013articulated}. As can be seen from the Table \ref{tab:loc}, our method outperforms existing techniques with various $\alpha$ setting in terms of PCK. In addition, the most striking result is that our approach outperforms the compared methods with large margins when using small $\alpha$ value. 

The part localization architecture adopted in DPS-CNN achieves a highest average \emph{PCK@0.1} $88.5\%$ on the CUB-200-2011 test set for $15$ object parts. Specifically, the employed Gaussian smoothing kernel delivers $2\%$ improvements over methods that use standard convolutional layers in BN-GoogleNet.

\begin{table}
\small
\centering
\caption{Receptive field size of different layers.}
\label{tab:rf}       

\begin{tabular}{cc}
\hline\noalign{\smallskip}
  Layer & Rec. Field \\
\noalign{\smallskip}\hline\noalign{\smallskip}
Inception-4a & $107\times107$ \\ 
Inception-4b & $139\times139$ \\ 
Inception-4c & $171\times171$ \\ 
Inception-4d & $204\times204$ \\ 
\noalign{\smallskip}\hline\noalign{\smallskip}
\end{tabular}
\end{table}

Another interesting phenomenon of note is that parts residing near the birds’ heads tend to be located more accurately. It turns out that a bird’s head has a relatively stable structure with fewer deformations and a lower probability of occlusion. In contrast, parts that are highly deformable such as the wings and legs get lower PCK values. Figure \ref{fig:loc} shows typical localization results using the proposed method.

\begin{table*}
\centering
\caption{Comparison of per-part PCK(\%) and over-all APK(\%) on CUB200-2011. The abbreviated part names from left to right are: Back, Beak, Belly, Breast, Crown, Forehead,Left Eye,Left Leg, Left Wing, Nape, Right Eye, Right Leg, Right Wing,Tail, and Throat}
\label{tab:loc}       
\begin{tabulary}{17.5cm}{cccccccccccccccccc|c}

\hline\noalign{\smallskip}

  $\alpha$ & Methods & Ba & Bk & Be & Br & Cr & Fh & Le & Ll & Lw & Na & Re & Rl & Rw & Ta & Th & Avg & APK  \\
\noalign{\smallskip}\hline\noalign{\smallskip}
\multirow{7}{*}{$0.1$} & \cite{huang2016part} & 80.7 & 89.4 & 79.4 & 79.9 & 89.4 & 88.5 & 85.0 & 75.0 & 67.0 & 85.7 & 86.1 & \textbf{77.5} & 67.8 & 76.0 & 90.8 & 81.2 & 86.6 \\
& \cite{zhang2015fine} & 85.6 & 94.9 & 81.9 & 84.5 & 94.8 & \textbf{96.0} & \textbf{95.7} & 64.6 & 67.8 & 90.7 & \textbf{93.8} & 64.9 & 69.3 & 74.7 & 94.5 & 83.6 & - \\
& \cite{yu2016deep} & \textbf{94.0} & 82.5 & \textbf{92.2} & 93.0 & 92.2 & 91.5 & 93.3 & 69.7 & 68.1 & 86.0 & \textbf{93.8} & 74.2 & 68.9 & 77.4 & 93.4 & 84.7 & - \\
& Ours(4a) & 82.7 & 94.1 & 85.3 & 87.8 & 95.2 & 93.3 & 88.6 & 75.5 & 75.9 & 92.0 & 89.5 & 76.6 & 75.9 & 67.4 & 94.7 & 84.9 & 89.1 \\
& Ours(4b) & 87.4 & 93.6 &  87.4 &  88.9 & 95.2 &  93.7 &  88.3 &   73.3 &  77.6 &   93.4 &  88.9 &   76.3 &   79.0 &   70.5 &   94.5  & 85.9 & 88.9 \\
& Ours(4c) & 89.0  &  \textbf{95.1}  &   91.5  &   92.6  &    \textbf{95.7}  &    94.7  &    90.3  &    \textbf{78.5}  &    \textbf{82.3} &    94.4  &    91.0  &    73.2  &    81.9  & 78.4 & 95.7 & 88.3 & 90.9 \\
& Ours(4d) & 89.0  &   95.0 &  92.2 &   \textbf{93.2} &    95.2 &   94.2 &   90.5 &   73.2 &   81.5 &   \textbf{94.4} &  91.6 &    75.5 &   \textbf{82.3} &   \textbf{83.2} &   \textbf{95.8}
 & \textbf{88.5} & \textbf{91.2} \\

\noalign{\smallskip}\hline
\multirow{7}{*}{$0.05$} & \cite{huang2016part} & 48.8 & 63.7 & 44.5 & 50.3 & 50.2 & 43.7 & 80.0 & 44.8 & 42.7 & 60.1 & 59.4 & 46.5 & 39.8 & 46.8 & 71.9 & 52.9 & 62.7 \\
& \cite{zhang2015fine} & 46.8 & 62.5 & 40.7 & 45.1 & 59.8 & 63.7 & 66.3 & 33.7 & 31.7 & 54.3 & 63.8 & 36.2 & 33.3 & 39.6 & 56.9 & 49.0 & - \\
& \cite{yu2016deep} & 66.4 & 49.2 & 56.4 & 60.4 & 61.0 & 60.0 & 66.9 & 32.3 & 35.8 & 53.1 & 66.3 & 35.0 & 37.1 & 40.9 & 65.9 & 52.4 & - \\
& Ours(4a) & \textbf{70.6} & \textbf{89.5} & \textbf{69.5} & \textbf{75.0} & \textbf{89.0} & \textbf{87.8} & \textbf{87.1} & \textbf{58.5} & \textbf{57.6} & \textbf{84.6} & \textbf{87.8} & \textbf{59.6} & \textbf{60.2} & \textbf{56.3} & \textbf{90.0} & \textbf{74.9} & \textbf{80.4} \\
& Ours(4b) & 69.2 &   79.4 &    69.0 &   74.5 &   73.2 &  72.3 &  85.7 &  53.3 &  58.3 &  83.7 & 86.0 &    55.5 &  60.1 &   59.0 &  86.5 & 74.5 & 71.1 \\
& Ours(4c) & 62.3 &  57.1 & 67.6 & 72.2 & 49.1 & 47.0 & 84.6 & 49.7 & 57.6 & 79.3 & 84.9 & 44.1 & 56.9 &   63.7 & 82.6 & 63.0 & 67.9 \\
& Ours(4d) & 42.3 & 27.5 & 59.7 & 60.6 &  21.3 &  23.3 & 82.2 &  33.1 & 49.6 &  65.6 & 82.4 &  37.4 & 47.5 &  66.7 &  69.4 & 51.3 & 54.5 \\
\noalign{\smallskip}\hline
\multirow{7}{*}{$0.02$} & \cite{huang2016part} & 11.1 & 16.9 & 9.1 & 11.2 & 5.2 & 4.1 & 40.4 & 9.4 & 10.8 & 14.6 & 9.9 & 11.9 & 9.6 & 11.2 & 22.3 & 13.2 & 13.3 \\
& \cite{zhang2015fine} & 9.4 & 12.7 & 8.2 & 12.2 & 13.2 & 11.3 & 7.8 & 6.7 & 11.5 & 12.5 & 7.3 & 6.2 & 8.2 & 11.8 & 56.9 & 13.1 & - \\
& \cite{yu2016deep} & 18.8 & 12.8 & 14.2 & 15.9 & 15.9 & 16.2 & 20.3 & 7.1 & 8.3 & 13.8 & 19.7 & 7.8 & 9.6 & 9.6 & 18.3 & 13.8 & - \\
& Ours(4a) & \textbf{24.9} & \textbf{31.0} & \textbf{23.0} & \textbf{28.3} & \textbf{25.1} & \textbf{26.6} & \textbf{44.8} & \textbf{19.6} & \textbf{17.4} & \textbf{38.4} & \textbf{46.9} & \textbf{20.9} & \textbf{20.7} & \textbf{22.0} & \textbf{37.5} & \textbf{28.5} & \textbf{17.2} \\
& Ours(4b) & 19.7 &  15.8 &  21.6 & 24.0 & 9.1 &   8.1 & 40.7 & 16.0 & 16.8 & 32.6 &  43.1 &   16.7 & 17.7 &   23.6 &  29.8 & 22.4 & 13.5 \\
& Ours(4c) & 12.5 &   5.9 & 17.9 &  17.9 &    2.6 &   3.0 & 41.4 &  12.0 &  15.0 & 22.2 &   41.4 &  8.9 & 14.9 &  24.0 &  23.1 & 17.5 & 11.8 \\
& Ours(4d) & 6.4 & 1.9 & 14.1 & 11.8 & 1.0 & 2.1 &  36.7 &    4.9 &  10.9 &  15.5 &  38.5 &    5.9 &  10.4 &  24.0 &  17.0 & 13.4 & 9.3 \\
\noalign{\smallskip}\hline
\end{tabulary}
\end{table*}

\begin{table*}
\centering
\caption{Localization recall of candidate points selected by \textit{inception-4a} layer with different $\alpha$ values. The abbreviated part names from left to right are: Back, Beak, Belly, Breast, Crown, Forehead,Left Eye,Left Leg, Left Wing, Nape, Right Eye, Right Leg, Right Wing, Tail, and Throat}
\label{tab:single}       

\begin{tabulary}{17.5cm}{cccccccccccccccccc}

\hline\noalign{\smallskip}

 Part & Ba & Bk & Be & Br & Cr & Fh & Le & Ll & Lw & Na & Re & Rl & Rw & Ta & Th \\
\noalign{\smallskip}\hline\noalign{\smallskip}
Accuracy(\%)&47.9 & 63.7 & 43.9 & 56.8 & 66.8 & 66.1 & 36.6 & 30.8 & 30.4 & 64.8 & 36.1 & 29.2 & 29.7 & 20.0 & 68.7 \\
 
\hline

\end{tabulary}
\end{table*}

\subsection{Classification results}

We begin our classification analysis by studying the discriminative power of each object part. We select one object part each time as the input and discard the computation of all other parts. As shown in Table \ref{tab:single}, different parts produce significantly different classification results. The most discriminative part "\emph{Throat}" achieves a quite impressive accuracy of $68.7\%$, while the lowest accuracy is $20.0\%$ for the part "\emph{Tail}". Therefore, to improve classification, it may be beneficial to find a rational combination or order of object parts instead of directly running the experiment on all parts altogether. 
More interestingly, when comparing the results between Table \ref{tab:loc} and Table \ref{tab:single} it can be seen that parts located more accurately such as  \emph{Throat}, \emph{Nape}, \emph{Forehead} and \emph{Beak} tend to achieve better performance in the recognition task, while some parts like \emph{Tail} and \emph{Left Leg} with poor localization accuracy perform worse. This observation may support the hypothesis that a more discriminative part is easier to locate in the context of fine-grained categorization and vice versa. 

To evaluate our framework’s overall performance, we first train a baseline model with accuracy $81.56\%$ using a BN-Inception architecture \cite{ioffe2015batch} with pre-training on ImageNet \cite{russakovsky2015imagenet}. By stacking certain part features and applying our proposed fusion method, our framework improves the performance to $85.12\%$. Also, to evaluate our proposed feature fusion method, we then train four DPS-CNN models with same experimental settings (maximum iteration and learning rate) but using different feature fusion methods. The results shown in Table \ref{tab:setting} (Rows 2-5) demonstrate that SMM fusion achieves the best performance and outperforms the FC method by $1.69\%$. 

To investigate which parts should be selected in our learning framework, we conduct the following experiments by employing two guiding principles: one concerns the feature discrimination and the other feature diversity. Here we consider parts with higher accuracy in Table  \ref{tab:single} are more discriminative, and combination of parts with distant location are more diverse. We firstly select top 6 parts with the highest accuracy from Table \ref{tab:single} by only applying the discriminative principle, then choose 3,5,9 and 15 parts respectively by taking two principles into account. Experimental results are shown in Table \ref{tab:setting} (Row 6-10), we observe that increasing part numbers generally bring slight improvement. However, all setting perform better than that with 6 most discriminative parts. This mainly because most of these parts are adjacent to each other so that it fails to produce diverse feature in our framework. Also, it should be noticed that using all parts feature does not guarantee the best performance, on the other hand, results in pool accuracy. This finding shows that the feature redundancy caused by appending exorbitant number of parts in learning, may degrade the accuracy, and suggests that an appropriate strategy for integrating multiple parts is critical.

\begin{table}
\small
\centering
\caption{Comparison of different settings of our approach  on CUB200-2011 .}
\label{tab:setting}       

\begin{tabular}{clc}
\hline\noalign{\smallskip}
  Row & Setting & Acc(\%)\\
\noalign{\smallskip}\hline\noalign{\smallskip}
1 & Object Only(Baseline) & 81.56 \\
\hline\noalign{\smallskip}
2 & 5-parts + FC  & 81.86  \\
3 & 5-parts + SS  & 83.06\\
4 & 5-parts + SM  & 83.41\\
5 & 5-parts + SMM & \textbf{83.55}  \\
\hline\noalign{\smallskip}
6 & 6-parts + SMM & 84.12  \\
\hline\noalign{\smallskip}
7 & 3-parts + SMM & 84.29  \\
8 & 5-parts + SMM & 84.91  \\
\textbf{9} & \textbf{9 parts + SMM} & \textbf{85.12}  \\
10 & 15-parts + SMM & 84.45  \\
\noalign{\smallskip}\hline\noalign{\smallskip}
\end{tabular}
\end{table}

\begin{table*}
\begin{center}
\begin{tabular}{l|cccc|c|c|c}
\hline
\multirow{2}{*}{Method} & \multicolumn{2}{c}{Training} & \multicolumn{2}{c|}{Testing} & \multirow{2}{*}{Pre-trained Model.}  & \multirow{2}{*}{FPS \footnote{Only FPS (Frames Per Second) results that has been reported by the authors are shown in this table.}}  & \multirow{2}{*}{Acc(\%)}.\\
\cline{2-5}
						& BBox		& Parts			   & BBox      & Parts            & & &\\
\hline
	Part-Stacked CNN \cite{huang2016part}     			& \checkmark & \checkmark & \checkmark   &   &  AlexNet 	& 20 & 76.62 \\
\hline
	Deep LAC \cite{lin2015deep}			      			& \checkmark & \checkmark & \checkmark   &   &  AlexNet     &  - & 80.26 \\
\hline
	Part R-CNN \cite{zhang2014part}		      			& \checkmark & \checkmark & \checkmark   &   &  AlexNet 	&  - & 76.37 \\
\hline
	SPDA-CNN \cite{zhangspda2016}               			& \checkmark & \checkmark & \checkmark 	 &   &  VGG16       &  - & 84.55 \\
\hline
	SPDA-CNN \cite{zhangspda2016}+ensemble               			& \checkmark & \checkmark & \checkmark 	 &   &  VGG16       &  - & 85.14 \\

\hline
	Part R-CNN \cite{zhang2014part} without BBox		      			& \checkmark & \checkmark &  			 &   &  AlexNet 	&  - & 73.89 \\
\hline
	PoseNorm CNN \cite{branson2014bird}	  	  			& \checkmark & \checkmark &  			 &   &  AlexNet	    &  - & 75.70 \\
\hline
	Bilinear-CNN (M+D+BBox) \cite{lin2015bilinear}	  			& \checkmark &  		  & \checkmark &   &  VGG16+VGGM	  &  8 & 85.10 \\
\hline
	Bilinear-CNN (M+D) \cite{lin2015bilinear}	  	&  			 &  		  &  			 &   &  VGG16+VGGM	  &  8 & 84.10 \\
\hline
	Constellation-CNN \cite{simon2015neural}	  			&  			 &            &  			 &   &  VGG19 	    &  - & 84.10 \\	
\hline 
	Spatial Transformer CNN \cite{jaderberg2015spatial}	&  			 &  		  &  			 &   & Inception+BN 	  &  - & 84.10 \\
\hline
	Two-Level \cite{xiao2015application}	  				&  			 &  		  &  			 &   &  VGG16 	    &  - & 77.90 \\
\hline
	Co-Segmentation \cite{krause2015fine}	 		    & \checkmark &  		  &  \checkmark &   &  VGG19	    &  - & 82.80 \\
\hline 
DPS-CNN with 9 parts  										&  			 & \checkmark &  			 &   & Inception+BN	  &  \textbf{32}   & 85.12 \\
\textbf{DPS-CNN ensemble with 4 models}   										&  			 & \checkmark &  			 &   & Inception+BN	  &  8   & \textbf{86.56} \\

\hline
\end{tabular}
\end{center} 
\caption{Comparison with state-of-the-art methods on the CUB-200-2011 dataset.  }
\label{tab:cls}
\end{table*}



We also present the performance comparison between DPS-CNN and existing fine-grained recognition methods. As can be seen in Table \ref{tab:cls}, our approach using only keypoint annotation during training achieve 85.12\% accuracy which is comparable with the state-of-the-art method \cite{lin2015bilinear} that achieves 85.10\% using bounding box both in training and testing. Moreover, it is interpretable and faster - the entire forward pass of DPS-CNN runs at $32$ \si{frames/sec} (NVIDIA TitanX), while B-CNN[D,M]\cite{lin2015bilinear} runs at $8$ \si{frames/sec} (NVIDIA K40)\footnote{note that the computational power of TitanX is around 1.5 times of that of K40).}. In particular, our method is much faster than  proposal based methods such as \cite{zhang2014part} and \cite{zhangspda2016} which require multiple network forward propagation for proposal evaluation, while part detection and feature extraction are accomplished efficiently by running one forward pass in our approach. In addition, we combine four models stemmed from integrating different parts(listed in Table \ref{tab:setting} (Row 7-10)) to form an ensemble which leads to 86.56\% accuracy on cub-200-2011.

To understand what features are learned in DPS-CNN, we use the aforementioned five-parts model and show its feature map visualization compared with that from BN-Inception model fine-tuning on cub-200-2011. Specifically, we pick the top six scoring feature maps of \emph{Inception-4a} layer for visualization, where the score is the sum over each feature map. As shown in Figure \ref{fig:fm}, each example image from test set is followed by three rows of feature maps, from top row to bottom, which are selected from part stream, object stream and BN-inception base-line network respectively. Interestingly, by comparison, our part stream have learned feature maps that appear to be more intuitive than those learned by the other two methods. Specifically, it yields more focused and cleaner patterns which tend to be highly activated by the network. Moreover, we can observe that  object stream and baseline network are more likely to activate filters with extremely high frequency details but at the expense of extra noise, while part stream tends to obtain a mixture of low and mid frequency information. The red dashed box in Figure \ref{fig:fm} indicates a failure example, in which both our part stream and object stream fails to learn useful feature. This may be caused by our part localization network fails to locate \textit{Crown} and \textit{Left Leg} parts because the branch in this image looks similar to bird legs and another occluded bird also has an effect on locating the \textit{Crown} part.

\begin{figure*}[t]
\begin{center}
\includegraphics[width=1\linewidth]{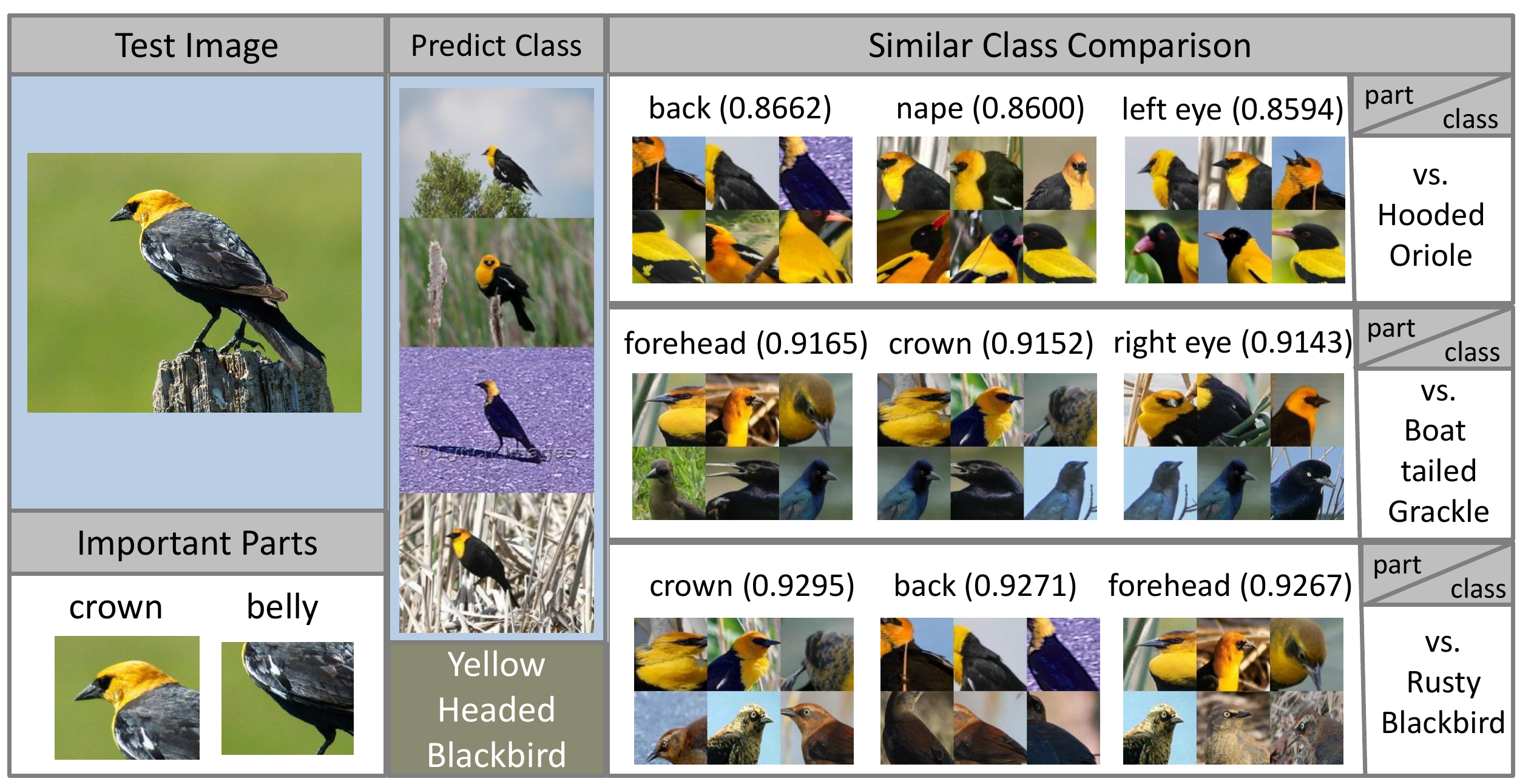}
\end{center}
   \caption{Example of the prediction manual generated by the proposed approach. Given a test image, the system reports its predicted class label with some typical exemplar images. Part-based comparison criteria between the predicted class and its most similar classes are shown in the right part of the image. The number in brackets shows the confidence of classifying two categories by introducing a specific part. We present top three object parts for each pair of comparison. For each of the parts, three part-center-cropped patches are shown for the predicted class (upper rows) and the compared class (lower rows) respectively.}
\label{fig:interpret}
\end{figure*}

\subsection{Model interpretation}
One of the most prominent features of DPS-CNN method is that it can produce human-understandable interpretation manuals for fine-grained recognition. Here we directly borrow the idea from \cite{huang2016part} for interpretation using the proposed method.

Different from \cite{berg2013poof} who directly conducted one-on-one classification on object parts, the interpretation process of the proposed method is conducted relatively indirectly. Since using each object part alone does not produce convincing classification results, we perform the interpretation analysis on a combination of bounding box supervision and each single object part. The analysis is performed in two ways: a "one-versus-rest" comparison to denote the most discriminative part to classify a subcategory from all other classes, and a "one-versus-one" comparison to obtain the classification criteria of a subcategory with its most similar classes.
\begin{itemize}
\item The \emph{``one-versus-rest''} manual for an object category $k$. For every part $p$, we compute the summation of prediction scores of the category's positive samples.
The most discriminative part is then captured as the one with the largest accumulated score:
\begin{equation}
p_k^* = \argmax_p \sum_{i, y_i=k}S^{(p)}_{ip}.
\end{equation}

\item The \emph{``one-versus-one''} manual obtained by computing as the part which results in the largest difference of prediction scores on two categories $k$ and $l$. We first take the respective two rows in the score matrix $S$, and re-normalize it using the binary classification criterion as $S'$. Afterwards, the most discriminative part is given as:
\begin{equation}
p_{k\rightarrow l}^* = \argmax_p (\sum_{i, y_i=k}S'^{(p)}_{ip}+\sum_{j, y_j=l}S'^{(p)}_{jp})
\end{equation}
\end{itemize}

The model interpretation routine is demonstrated in Figure \ref{fig:interpret}. When a test image is presented, the proposed method first conducts object classification using the DPS-CNN architecture. The predicted category is presented as a set of images in the dataset that are closest to the test image according to the feature vector of each part. Except for the classification results, the proposed method also presents classification criteria that distinguish the predicted category from its most similar neighboring classes based on object parts. Again we use part features but after part cropping to retrieve nearest neighbor part patches of the input test image. The procedure described above provides an intuitive visual guide for distinguishing fine-grained categories.

\section{Conclusion}\label{sec:conclusion}

In this paper, we propose a novel fine-grained recognition method called Deeper Part-Stacked CNN (DPS-CNN). The method exploits detailed part-level supervision, in which object parts are first located by a localization network and then by a two-stream classification network that explicitly captures object- and part-level information. We also present a new feature vector fusion strategy that effectively combines both part and object stream features. Experiments on CUB-200-2011 demonstrate the effectiveness and efficiency of our system. We also present human-understandable interpretations of the proposed method, which can be used as a visual field guide for studying fine-grained categorization. 

DPS-CNN can be applied to fine-grained visual categorization with strong supervision and can be easily generalized to various applications including: 

\begin{enumerate}

\item Discarding the requirement for strong supervision. Instead of introducing manually labeled part annotations to generate human-understandable visual guides, one can also exploit unsupervised part discovery methods \cite{krause2015fine} to define object parts automatically, which requires far less human labelling effort.
\item Attribute learning. The application of DPS-CNN is not restricted to FGVC. For instance, online shopping \cite{kiapour2015where} performance could benefit from clothing attribute analysis from local parts provided by DPS-CNN. 
\item Context-based CNN. The role of local “parts” in DPS-CNN is interchangeable with global contexts, especially for objects that are small and have no obvious object parts such as volleyballs or tennis balls.
\end{enumerate}

\ifCLASSOPTIONcaptionsoff
  \newpage
\fi



%
\bibliographystyle{IEEEtran}
\bibliography{egbib}

\begin{thebibliography}{10}
\providecommand{\url}[1]{#1}
\csname url@samestyle\endcsname
\providecommand{\newblock}{\relax}
\providecommand{\bibinfo}[2]{#2}
\providecommand{\BIBentrySTDinterwordspacing}{\spaceskip=0pt\relax}
\providecommand{\BIBentryALTinterwordstretchfactor}{4}
\providecommand{\BIBentryALTinterwordspacing}{\spaceskip=\fontdimen2\font plus
\BIBentryALTinterwordstretchfactor\fontdimen3\font minus
  \fontdimen4\font\relax}
\providecommand{\BIBforeignlanguage}[2]{{%
\expandafter\ifx\csname l@#1\endcsname\relax
\typeout{** WARNING: IEEEtran.bst: No hyphenation pattern has been}%
\typeout{** loaded for the language `#1'. Using the pattern for}%
\typeout{** the default language instead.}%
\else
\language=\csname l@#1\endcsname
\fi
#2}}
\providecommand{\BIBdecl}{\relax}
\BIBdecl

\bibitem{welinder2010caltech}
P.~Welinder, S.~Branson, T.~Mita, C.~Wah, F.~Schroff, S.~Belongie, and
  P.~Perona, ``Caltech-ucsd birds 200,'' 2010.

\bibitem{wah2011caltech}
C.~Wah, S.~Branson, P.~Welinder, P.~Perona, and S.~Belongie, ``The caltech-ucsd
  birds-200-2011 dataset,'' 2011.

\bibitem{berg2014birdsnap}
T.~Berg, J.~Liu, S.~W. Lee, M.~L. Alexander, D.~W. Jacobs, and P.~N. Belhumeur,
  ``Birdsnap: Large-scale fine-grained visual categorization of birds,'' in
  \emph{Computer Vision and Pattern Recognition (CVPR), 2014}.\hskip 1em plus
  0.5em minus 0.4em\relax IEEE, 2014, pp. 2019--2026.

\bibitem{khosla2011novel}
A.~Khosla, N.~Jayadevaprakash, B.~Yao, and F.-F. Li, ``Novel dataset for
  fine-grained image categorization: Stanford dogs,'' in \emph{Proc. CVPR
  Workshop on Fine-Grained Visual Categorization (FGVC)}, 2011.

\bibitem{parkhi2012cats}
O.~M. Parkhi, A.~Vedaldi, A.~Zisserman, and C.~Jawahar, ``Cats and dogs,'' in
  \emph{Computer Vision and Pattern Recognition (CVPR), 2012 IEEE Conference
  on}.\hskip 1em plus 0.5em minus 0.4em\relax IEEE, 2012, pp. 3498--3505.

\bibitem{nilsback2008automated}
M.-E. Nilsback and A.~Zisserman, ``Automated flower classification over a large
  number of classes,'' in \emph{Computer Vision, Graphics \& Image Processing,
  2008. ICVGIP'08. Sixth Indian Conference on}.\hskip 1em plus 0.5em minus
  0.4em\relax IEEE, 2008, pp. 722--729.

\bibitem{angelova2013image}
A.~Angelova, S.~Zhu, and Y.~Lin, ``Image segmentation for large-scale
  subcategory flower recognition,'' in \emph{Applications of Computer Vision
  (WACV), 2013 IEEE Workshop on}.\hskip 1em plus 0.5em minus 0.4em\relax IEEE,
  2013, pp. 39--45.

\bibitem{stark2011fine}
M.~Stark, J.~Krause, B.~Pepik, D.~Meger, J.~J. Little, B.~Schiele, and
  D.~Koller, ``Fine-grained categorization for 3d scene understanding,''
  \emph{International Journal of Robotics Research}, vol.~30, no.~13, pp.
  1543--1552, 2011.

\bibitem{maji2013fine}
S.~Maji, E.~Rahtu, J.~Kannala, M.~Blaschko, and A.~Vedaldi, ``Fine-grained
  visual classification of aircraft,'' \emph{arXiv preprint arXiv:1306.5151},
  2013.

\bibitem{wah2011multiclass}
C.~Wah, S.~Branson, P.~Perona, and S.~Belongie, ``Multiclass recognition and
  part localization with humans in the loop,'' in \emph{Computer Vision (ICCV),
  2011 IEEE International Conference on}.\hskip 1em plus 0.5em minus
  0.4em\relax IEEE, 2011, pp. 2524--2531.

\bibitem{vedaldi2014understanding}
A.~Vedaldi, S.~Mahendran, S.~Tsogkas, S.~Maji, R.~Girshick, J.~Kannala,
  E.~Rahtu, I.~Kokkinos, M.~B. Blaschko, D.~Weiss \emph{et~al.},
  ``Understanding objects in detail with fine-grained attributes,'' in
  \emph{Computer Vision and Pattern Recognition (CVPR), 2014 IEEE Conference
  on}.\hskip 1em plus 0.5em minus 0.4em\relax IEEE, 2014, pp. 3622--3629.

\bibitem{krause2015fine}
J.~Krause, H.~Jin, J.~Yang, and L.~Fei-Fei, ``Fine-grained recognition without
  part annotations,'' in \emph{Proceedings of the IEEE Conference on Computer
  Vision and Pattern Recognition}, 2015, pp. 5546--5555.

\bibitem{xu2015augmenting}
Z.~Xu, S.~Huang, Y.~Zhang, and D.~Tao, ``{Augmenting strong supervision using
  web data for fine-grained categorization},'' in \emph{Computer Vision (ICCV),
  2015 IEEE International Conference on}, 2015.

\bibitem{deng2013fine}
J.~Deng, J.~Krause, and L.~Fei-Fei, ``Fine-grained crowdsourcing for
  fine-grained recognition,'' in \emph{Computer Vision and Pattern Recognition
  (CVPR), 2013 IEEE Conference on}.\hskip 1em plus 0.5em minus 0.4em\relax
  IEEE, 2013, pp. 580--587.

\bibitem{chai2013symbiotic}
Y.~Chai, V.~Lempitsky, and A.~Zisserman, ``Symbiotic segmentation and part
  localization for fine-grained categorization,'' in \emph{Proceedings of the
  IEEE International Conference on Computer Vision}, 2013, pp. 321--328.

\bibitem{branson2014bird}
S.~Branson, G.~Van~Horn, S.~Belongie, and P.~Perona, ``Bird species
  categorization using pose normalized deep convolutional nets,'' \emph{arXiv
  preprint arXiv:1406.2952}, 2014.

\bibitem{lin2015bilinear}
T.-Y. Lin, A.~RoyChowdhury, and S.~Maji, ``Bilinear cnn models for fine-grained
  visual recognition,'' in \emph{Proceedings of the IEEE International
  Conference on Computer Vision}, 2015, pp. 1449--1457.

\bibitem{wang2015multiple}
D.~Wang, Z.~Shen, J.~Shao, W.~Zhang, X.~Xue, and Z.~Zhang, ``Multiple
  granularity descriptors for fine-grained categorization,'' in
  \emph{Proceedings of the IEEE International Conference on Computer Vision},
  2015, pp. 2399--2406.

\bibitem{berg2013poof}
T.~Berg and P.~Belhumeur, ``Poof: Part-based one-vs.-one features for
  fine-grained categorization, face verification, and attribute estimation,''
  in \emph{Proceedings of the IEEE Conference on Computer Vision and Pattern
  Recognition}, 2013, pp. 955--962.

\bibitem{berg2013you}
T.~Berg and P.~N. Belhumeur, ``How do you tell a blackbird from a crow?'' in
  \emph{Computer Vision (ICCV), 2013 IEEE International Conference on}.\hskip
  1em plus 0.5em minus 0.4em\relax IEEE, 2013, pp. 9--16.

\bibitem{kumar2012leafsnap}
N.~Kumar, P.~N. Belhumeur, A.~Biswas, D.~W. Jacobs, W.~J. Kress, I.~C. Lopez,
  and J.~V. Soares, ``Leafsnap: A computer vision system for automatic plant
  species identification,'' in \emph{Computer Vision--ECCV 2012}.\hskip 1em
  plus 0.5em minus 0.4em\relax Springer, 2012, pp. 502--516.

\bibitem{branson2014ignorant}
S.~Branson, G.~Van~Horn, C.~Wah, P.~Perona, and S.~Belongie, ``The ignorant led
  by the blind: A hybrid human--machine vision system for fine-grained
  categorization,'' \emph{International Journal of Computer Vision}, vol. 108,
  no. 1-2, pp. 3--29, 2014.

\bibitem{van2015building}
G.~Van~Horn, S.~Branson, R.~Farrell, S.~Haber, J.~Barry, P.~Ipeirotis,
  P.~Perona, and S.~Belongie, ``Building a bird recognition app and large scale
  dataset with citizen scientists: The fine print in fine-grained dataset
  collection,'' in \emph{Proceedings of the IEEE Conference on Computer Vision
  and Pattern Recognition}, 2015, pp. 595--604.

\bibitem{krizhevsky2012imagenet}
A.~Krizhevsky, I.~Sutskever, and G.~E. Hinton, ``Imagenet classification with
  deep convolutional neural networks,'' in \emph{Advances in neural information
  processing systems}, 2012, pp. 1097--1105.

\bibitem{rosch1976basic}
E.~Rosch, C.~B. Mervis, W.~D. Gray, D.~M. Johnson, and P.~Boyes-Braem, ``Basic
  objects in natural categories,'' \emph{Cognitive psychology}, vol.~8, no.~3,
  pp. 382--439, 1976.

\bibitem{maji2014part}
S.~Maji and G.~Shakhnarovich, ``Part and attribute discovery from relative
  annotations,'' \emph{International Journal of Computer Vision}, vol. 108, no.
  1-2, pp. 82--96, 2014.

\bibitem{zhang2014part}
N.~Zhang, J.~Donahue, R.~Girshick, and T.~Darrell, ``Part-based r-cnns for
  fine-grained category detection,'' in \emph{Computer Vision--ECCV
  2014}.\hskip 1em plus 0.5em minus 0.4em\relax Springer, 2014, pp. 834--849.

\bibitem{zhang2014fused}
X.~Zhang, H.~Xiong, W.~Zhou, and Q.~Tian, ``Fused one-vs-all mid-level features
  for fine-grained visual categorization,'' in \emph{Proceedings of the ACM
  International Conference on Multimedia}.\hskip 1em plus 0.5em minus
  0.4em\relax ACM, 2014, pp. 287--296.

\bibitem{ioffe2015batch}
S.~Ioffe and C.~Szegedy, ``Batch normalization: Accelerating deep network
  training by reducing internal covariate shift,'' in \emph{Proceedings of the
  32nd International Conference on Machine Learning (ICML-15)}, 2015, pp.
  448--456.

\bibitem{huang2016part}
S.~Huang, Z.~Xu, D.~Tao, and Y.~Zhang, ``Part-stacked cnn for fine-grained
  visual categorization,'' in \emph{Proceedings of the IEEE International
  Conference on Computer Vision}, 2016.

\bibitem{zhang2015fine}
N.~Zhang, E.~Shelhamer, Y.~Gao, and T.~Darrell, ``Fine-grained pose prediction,
  normalization, and recognition,'' \emph{arXiv preprint arXiv:1511.07063},
  2015.

\bibitem{zhangspda2016}
H.~Zhang, T.~Xu, M.~Elhoseiny, X.~Huang, S.~Zhang, A.~Elgammal, and D.~Metaxas,
  ``Spda-cnn: Unifying semantic part detection and abstraction for fine-grained
  recognition.''

\bibitem{zhang2016picking}
X.~Zhang, H.~Xiong, W.~Zhou, W.~Lin, and Q.~Tian, ``Picking deep filter
  responses for fine-grained image recognition,'' in \emph{Proceedings of the
  IEEE Conference on Computer Vision and Pattern Recognition}, 2016, pp.
  1134--1142.

\bibitem{zhang2014panda}
N.~Zhang, M.~Paluri, M.~Ranzato, T.~Darrell, and L.~Bourdev, ``Panda: Pose
  aligned networks for deep attribute modeling,'' in \emph{Computer Vision and
  Pattern Recognition (CVPR), 2014 IEEE Conference on}.\hskip 1em plus 0.5em
  minus 0.4em\relax IEEE, 2014, pp. 1637--1644.

\bibitem{gkioxari2015actions}
G.~Gkioxari, R.~Girshick, and J.~Malik, ``Actions and attributes from wholes
  and parts,'' in \emph{Proceedings of the IEEE International Conference on
  Computer Vision}, 2015, pp. 2470--2478.

\bibitem{zhu2015deepm}
J.~Zhu, X.~Chen, and A.~L. Yuille, ``Deepm: A deep part-based model for object
  detection and semantic part localization,'' \emph{arXiv preprint
  arXiv:1511.07131}, 2015.

\bibitem{milborrow2008locating}
S.~Milborrow and F.~Nicolls, ``Locating facial features with an extended active
  shape model,'' in \emph{European conference on computer vision}.\hskip 1em
  plus 0.5em minus 0.4em\relax Springer, 2008, pp. 504--513.

\bibitem{cootes2001active}
T.~F. Cootes, G.~J. Edwards, C.~J. Taylor \emph{et~al.}, ``Active appearance
  models,'' \emph{IEEE Transactions on pattern analysis and machine
  intelligence}, vol.~23, no.~6, pp. 681--685, 2001.

\bibitem{matthews2004active}
I.~Matthews and S.~Baker, ``Active appearance models revisited,''
  \emph{International Journal of Computer Vision}, vol.~60, no.~2, pp.
  135--164, 2004.

\bibitem{saragih2009face}
J.~M. Saragih, S.~Lucey, and J.~F. Cohn, ``Face alignment through subspace
  constrained mean-shifts,'' in \emph{2009 IEEE 12th International Conference
  on Computer Vision}.\hskip 1em plus 0.5em minus 0.4em\relax IEEE, 2009, pp.
  1034--1041.

\bibitem{liu2013bird}
J.~Liu and P.~N. Belhumeur, ``Bird part localization using exemplar-based
  models with enforced pose and subcategory consistency,'' in \emph{Computer
  Vision (ICCV), 2013 IEEE International Conference on}.\hskip 1em plus 0.5em
  minus 0.4em\relax IEEE, 2013, pp. 2520--2527.

\bibitem{liu2014part}
J.~Liu, Y.~Li, and P.~N. Belhumeur, ``Part-pair representation for part
  localization,'' in \emph{Computer Vision--ECCV 2014}.\hskip 1em plus 0.5em
  minus 0.4em\relax Springer, 2014, pp. 456--471.

\bibitem{shih2015part}
K.~J. Shih, A.~Mallya, S.~Singh, and D.~Hoiem, ``Part localization using
  multi-proposal consensus for fine-grained categorization,'' \emph{Proceedings
  of The British Machine Vision Conference (BMVC)}, 2015.

\bibitem{lin2015deep}
D.~Lin, X.~Shen, C.~Lu, and J.~Jia, ``Deep lac: Deep localization, alignment
  and classification for fine-grained recognition,'' in \emph{Proceedings of
  the IEEE Conference on Computer Vision and Pattern Recognition}, 2015, pp.
  1666--1674.

\bibitem{yu2016deep}
X.~Yu, F.~Zhou, and M.~Chandraker, ``Deep deformation network for object
  landmark localization,'' \emph{arXiv preprint arXiv:1605.01014}, 2016.

\bibitem{girshick2015fast}
R.~Girshick, ``Fast r-cnn,'' in \emph{Proceedings of the IEEE International
  Conference on Computer Vision}, 2015, pp. 1440--1448.

\bibitem{long2015fully}
J.~Long, E.~Shelhamer, and T.~Darrell, ``Fully convolutional networks for
  semantic segmentation,'' in \emph{Proceedings of the IEEE Conference on
  Computer Vision and Pattern Recognition}, 2015, pp. 3431--3440.

\bibitem{wei2016convolutional}
S.-E. Wei, V.~Ramakrishna, T.~Kanade, and Y.~Sheikh, ``Convolutional pose
  machines,'' 2016.

\bibitem{bo2010kernel}
L.~Bo, X.~Ren, and D.~Fox, ``Kernel descriptors for visual recognition,'' in
  \emph{Advances in neural information processing systems}, 2010, pp. 244--252.

\bibitem{sanchez2011fisher}
J.~S{\'a}nchez, F.~Perronnin, and Z.~Akata, ``Fisher vectors for fine-grained
  visual categorization,'' in \emph{FGVC Workshop in IEEE Computer Vision and
  Pattern Recognition (CVPR)}, 2011.

\bibitem{szegedy2014going}
C.~Szegedy, W.~Liu, Y.~Jia, P.~Sermanet, S.~Reed, D.~Anguelov, D.~Erhan,
  V.~Vanhoucke, and A.~Rabinovich, ``Going deeper with convolutions,''
  \emph{arXiv preprint arXiv:1409.4842}, 2014.

\bibitem{simonyan2014very}
K.~Simonyan and A.~Zisserman, ``Very deep convolutional networks for
  large-scale image recognition,'' \emph{arXiv preprint arXiv:1409.1556}, 2014.

\bibitem{gavves2015local}
E.~Gavves, B.~Fernando, C.~G. Snoek, A.~W. Smeulders, and T.~Tuytelaars,
  ``Local alignments for fine-grained categorization,'' \emph{International
  Journal of Computer Vision}, vol. 111, no.~2, pp. 191--212, 2015.

\bibitem{branson2010visual}
S.~Branson, C.~Wah, F.~Schroff, B.~Babenko, P.~Welinder, P.~Perona, and
  S.~Belongie, ``Visual recognition with humans in the loop,'' in
  \emph{Computer Vision--ECCV 2010}.\hskip 1em plus 0.5em minus 0.4em\relax
  Springer, 2010, pp. 438--451.

\bibitem{wah2014similarity}
C.~Wah, G.~Van~Horn, S.~Branson, S.~Maji, P.~Perona, and S.~Belongie,
  ``Similarity comparisons for interactive fine-grained categorization,'' in
  \emph{Computer Vision and Pattern Recognition (CVPR), 2014 IEEE Conference
  on}.\hskip 1em plus 0.5em minus 0.4em\relax IEEE, 2014, pp. 859--866.

\bibitem{simon2015neural}
M.~Simon and E.~Rodner, ``Neural activation constellations: Unsupervised part
  model discovery with convolutional networks,'' in \emph{Proceedings of the
  IEEE International Conference on Computer Vision}, 2015, pp. 1143--1151.

\bibitem{jia2014caffe}
Y.~Jia, E.~Shelhamer, J.~Donahue, S.~Karayev, J.~Long, R.~Girshick,
  S.~Guadarrama, and T.~Darrell, ``Caffe: Convolutional architecture for fast
  feature embedding,'' in \emph{Proceedings of the ACM International Conference
  on Multimedia}.\hskip 1em plus 0.5em minus 0.4em\relax ACM, 2014, pp.
  675--678.

\bibitem{ren2015faster}
S.~Ren, K.~He, R.~Girshick, and J.~Sun, ``Faster r-cnn: Towards real-time
  object detection with region proposal networks,'' \emph{arXiv preprint
  arXiv:1506.01497}, 2015.

\bibitem{uijlings2013selective}
J.~R. Uijlings, K.~E. van~de Sande, T.~Gevers, and A.~W. Smeulders, ``Selective
  search for object recognition,'' \emph{International journal of computer
  vision}, vol. 104, no.~2, pp. 154--171, 2013.

\bibitem{felzenszwalb2010object}
P.~F. Felzenszwalb, R.~B. Girshick, D.~McAllester, and D.~Ramanan, ``Object
  detection with discriminatively trained part-based models,'' \emph{Pattern
  Analysis and Machine Intelligence, IEEE Transactions on}, vol.~32, no.~9, pp.
  1627--1645, 2010.

\bibitem{yang2013articulated}
Y.~Yang and D.~Ramanan, ``Articulated human detection with flexible mixtures of
  parts,'' \emph{Pattern Analysis and Machine Intelligence, IEEE Transactions
  on}, vol.~35, no.~12, pp. 2878--2890, 2013.

\bibitem{long2014convnets}
J.~L. Long, N.~Zhang, and T.~Darrell, ``Do convnets learn correspondence?'' in
  \emph{Advances in Neural Information Processing Systems}, 2014, pp.
  1601--1609.

\bibitem{russakovsky2015imagenet}
O.~Russakovsky, J.~Deng, H.~Su, J.~Krause, S.~Satheesh, S.~Ma, Z.~Huang,
  A.~Karpathy, A.~Khosla, M.~Bernstein \emph{et~al.}, ``Imagenet large scale
  visual recognition challenge,'' \emph{International Journal of Computer
  Vision}, vol. 115, no.~3, pp. 211--252, 2015.

\bibitem{jaderberg2015spatial}
M.~Jaderberg, K.~Simonyan, A.~Zisserman \emph{et~al.}, ``Spatial transformer
  networks,'' in \emph{Advances in Neural Information Processing Systems},
  2015, pp. 2017--2025.

\bibitem{xiao2015application}
T.~Xiao, Y.~Xu, K.~Yang, J.~Zhang, Y.~Peng, and Z.~Zhang, ``The application of
  two-level attention models in deep convolutional neural network for
  fine-grained image classification,'' in \emph{Proceedings of the IEEE
  Conference on Computer Vision and Pattern Recognition}, 2015, pp. 842--850.

\bibitem{kiapour2015where}
K.~M.~Hadi, H.~Xufeng, L.~Svetlana, B.~Alexander, and B.~Tamara, ``Where to buy
  it: Matching street clothing photos in online shops,'' in \emph{Computer
  Vision (ICCV), 2015 IEEE International Conference on}, 2015.

\end{thebibliography}
\end{document}